%% file: main.tex
\documentclass[]{phai}

\input{math_commands.tex}

\usepackage{pifont}
\usepackage{makecell}
\usepackage{adjustbox}
\usepackage{array}
\usepackage{amssymb}
\usepackage{colortbl}
\usepackage{algorithm}
\usepackage{algpseudocode}
\usepackage{mathtools}

\newcommand{\cmark}{\ding{51}}

\title{OpenLongTail: Generative Scaling \\ of Long-Tail Driving Data}

\author[*1]{Lulin Liu}
\author[*1]{Nuo Chen}
\author[2]{Yan Wang}
\author[3]{Bangya Liu}
\author[4]{Wenyan Cong}
\author[4]{Hezhen Hu}
\author[2]{Boris Ivanovic}
\author[1]{Hao Wang}
\author[5]{Ziyao Zeng}
\author[6]{Xinyu Gong}
\author[1]{Yang Zhou}
\author[1]{Zixiang Xiong}
\author[7]{Dilin Wang}
\author[4]{Zhangyang Wang}
\author[8]{Weisong Shi}
\author[9]{Ruohan Zhang}
\author[2,9]{Marco Pavone}
\author[1,\dagger]{Zhiwen Fan}

\affiliation[1]{Texas A\&M University}
\affiliation[2]{NVIDIA}
\affiliation[3]{UW--Madison}
\affiliation[4]{UT Austin}
\affiliation[5]{Yale University}
\affiliation[6]{Adobe}
\affiliation[7]{Meta}
\affiliation[8]{University of Delaware}
\affiliation[9]{Stanford University}

\contribution[*]{Equal contribution}
\contribution[\dagger]{Corresponding author}

\abstract{
\input{sections/01_abstract}
}

\date{\today}

\metadata[Keywords]{Autonomous Driving, Video Diffusion Model, VLA Model}
\metadata[Project Website]{\url{https://openlongtail.github.io/}}

\begin{document}

\maketitle


\input{sections/02_introduction}
\input{sections/03_related_work}
\input{sections/04_problem_statement}
\input{sections/05_methodology}
\input{sections/06_evaluation}

\input{sections/07_conclusion}

\bibliographystyle{assets/plainnat}
\bibliography{example}

\clearpage
\appendix
\input{sections/08_appendix}

\end{document}

%% file: math_commands.tex
\usepackage{amsmath,amsfonts,bm}

\def\eqref#1{equation~\ref{#1}}

\def\1{\bm{1}}

\DeclareMathAlphabet{\mathsfit}{\encodingdefault}{\sfdefault}{m}{sl}
\SetMathAlphabet{\mathsfit}{bold}{\encodingdefault}{\sfdefault}{bx}{n}



%% file: sections/01_abstract.tex
Scaling robust driving policies is fundamentally bottlenecked by the scarcity of edge cases in curated datasets.
While the real world continuously captures these critical events, such long-tail events remain underutilized when collected from heterogeneous sources. Specifically, diverse but valuable in-the-wild long-tail videos lack the full view coverage required for training policy models, often missing multi-view poses or originating solely from monocular dash cameras.
This modality gap prevents these ubiquitous observations from being converted into scalable training data for long-tail generalization.
We introduce \textbf{\textit{OpenLongTail}}, an open-source generative data engine for scaling autonomous driving policies under long-tail events. To transform heterogeneous data sources into view-aligned and temporally coherent multi-view assets that are useful for policy learning, we develop a pose-informed extrapolative view synthesis pipeline that generates the missing views. We further enhance cross-view consistency and the temporal alignment for the newly generated views by injecting Plücker ray geometry into the scalable generation engine.
By synthesizing heterogeneous long-tail data, we observe a significant improvement in closed-loop driving robustness in handling long-tail events. By measuring the extrapolative view synthesis and pose metrics, we validate the effectiveness of OpenLongTail in visual fidelity, cross-view consistency, and ego-trajectory recovery.

%% file: sections/02_introduction.tex
\section{Introduction}
\label{sec:introduction}

Empowered by large-scale driving datasets~\citep{sun2020scalability,ettinger2021large,caesar2020nuscenes,xu2025wod}, recent  Vision-Language-Action (VLA) driving policies have made substantial progress on common scenarios by learning end-to-end mappings from scene understanding to vehicle control under well-structured training conditions~\citep{wang2025alpamayo,zhou2025autovla,zhou2026opendrivevla,xu2024vlm,yuan2025autodrive, jiang2025alphadrive,hwang2024emma,jiang2024senna,feng2025verdi,jiang2025diffvla}. However, real-world autonomy also requires these policies to remain reliable in long-tail situations, where dynamic obstacles and atypical environmental conditions might lead to safety-critical failures. Yet several challenges arise for the training data. First, events like animals on the road and work zones are scarce in curated datasets and are usually more expensive to capture at scale with calibrated multi-camera rigs. Sometimes though such events are recorded, the resulting data is fragmented across heterogeneous sources, spanning from recently released large-scale repositories such as the NVIDIA PhysicalAI Autonomous Vehicles (PAV)~\citep{nvidia_physicalai_av_2025} and Waymo E2E dataset \citep{xu2025wod} to widely available monocular dash-camera videos. These observations often follow incompatible annotation formats and provide missing or unreliable metric camera poses. This modality gap prevents common real-world long-tail videos from being directly converted into synchronized multi-view training assets, limiting the continued scaling of learned driving policies toward robust long-tail generalization.

\begin{figure}[t]
  \centering
  \includegraphics[width=\linewidth]{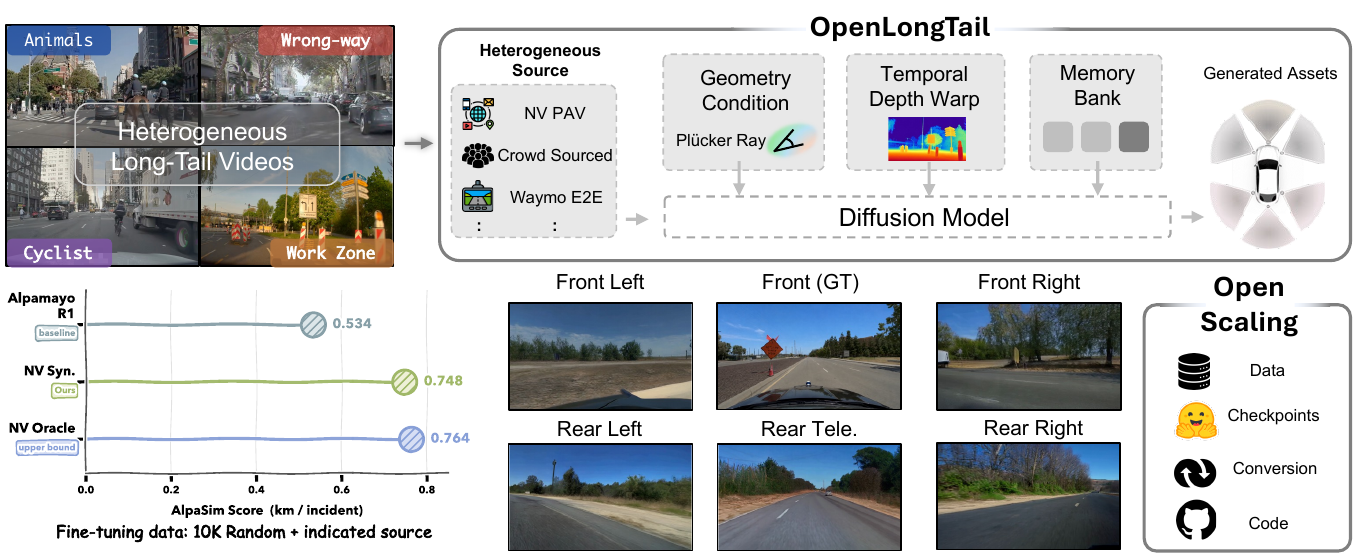}
\caption{\textbf{OpenLongTail} transforms heterogeneous long-tail driving videos from diverse sources into pose-grounded, synchronized multi-view assets under a target camera rig. It conditions a diffusion model on Pl\"ucker-ray geometry, temporal depth warping, and cross-view memory to synthesize extrapolative views beyond the observed front camera. The lower-left plot shows that fine-tuning with our synthesized long-tail data substantially improves closed-loop driving performance and approaches training with ground-truth multi-view data. To support open scaling, we release the generated data, model checkpoints, conversion tools, and code.}
  \label{fig:teaser}
  \vspace{-4mm}

\end{figure}

Bridging this gap requires a conversion engine that turns unposed or monocular driving data into synchronized, pose-grounded assets compatible with state-of-the-art models~\citep{zhou2025autovla,zhou2026opendrivevla,gao2026steervla}. Existing approaches fall short because the target output is a target-rig rollout whose side and rear cameras may observe regions absent from the input video. While recent video generation and camera-controlled synthesis methods have made strong progress in producing high-quality sequences under controllable poses~\citep{yu2025trajectorycrafter,bai2025recammaster,ren2025gen3c,lin2026vista4d,van2026anyview,yu2024viewcrafter,wu2025cat4d,yang2026neoverse}, they struggle when extrapolating into unobserved spaces. Specifically, they usually either lack cross-view consistency or rely on conditions like 3D bounding boxes and LiDAR, which are absent in heterogeneous videos~\citep{gao2023magicdrive,wang2024drivedreamer,zhao2025drivedreamer,ren2025cosmos,hu2023gaia,russell2025gaia, wang2024driving}. Explicit reconstruction with neural rendering (e.g., 3D Gaussian Splatting, 3DGS \citep{kerbl20233d}) faces the limitation where those methods depend on sufficient view overlap and cross-view photometric consistency, making them brittle under narrow monocular coverage, dynamic objects, and capturing artifacts. 3DGS conversion also depends on accurate camera trajectories, which are hard to acquire from the video sequences in the wild.

To bridge this gap, we introduce \textbf{OpenLongTail} (Figure~\ref{fig:teaser}), an open-source generative data engine that unifies heterogeneous long-tail driving videos into spatially aligned and temporally coherent multi-view assets for safer policy learning.  OpenLongTail utilizes the generalization capability from GFMs to recover a metric-scale ego-trajectory from weakly annotated driving videos, then turns this recovered motion into geometry conditions that guide pose-informed extrapolative view synthesis via a generation engine. The generative diffusion model injects Plücker ray geometry to encode target-camera rays and uses temporal depth warping to propagate visible evidence from the observed front view into the target rig, enabling missing side and rear contexts to be synthesized under a shared camera frame. 
Ultimately, OpenLongTail transforms the dash-cam driving logs into robust, posed multi-view assets. Through pose-recovery benchmarks, extrapolative view synthesis, and closed-loop policy evaluation, we demonstrate that OpenLongTail significantly improves driving robustness during long-tail events and enhances visual fidelity and ego-trajectory accuracy.
OpenLongTail introduces these major contributions: \vspace{-5pt}
\begin{enumerate}
\item We introduce \textbf{OpenLongTail}, an open-scaling generative data engine that converts heterogeneous long-tail driving videos into pose-grounded, multi-view data, enabling scalable VLA policy learning from heterogeneous sources.

\item We integrate Plücker-ray camera geometry and a cross-view memory bank into the generation process, allowing synthesized target views to share spatially grounded latent context and achieve stronger cross-view consistency under the target camera rig.

\item We validate OpenLongTail with downstream closed-loop VLA evaluation in AlpaSim, along with view synthesis and ego-trajectory recovery, showing that our generated assets significantly improve long-tail driving policy robustness, effectively unlocking ubiquitous monocular long-tail videos as a scalable training source.



\end{enumerate}







%% file: sections/03_related_work.tex
\section{Related Work}
\label{sec:related_work}

\paragraph{Vision-Language-Action Models For Autonomous Driving.}
Recent vision-language-action (VLA) models unify perception, language-conditioned reasoning, and end-to-end planning \citep{wang2025alpamayo,zhou2025autovla,zhou2026opendrivevla,li2025drivevla,sima2024drivelm,shao2024lmdrive,xu2024drivegpt4,wang2023drivemlm,ma2024dolphins}, increasingly leveraging reinforcement learning, diffusion, and explainable policies \citep{jiang2025alphadrive,hwang2024emma,jiang2024senna,feng2025verdi,jiang2025diffvla,rowe2025poutine,song2025lmad}. Despite success in nominal conditions, scaling these policies to rare, long-tail events remains bottlenecked by a strict reliance on curated, synchronized multi-view logs. Standard benchmarks provide rich sensor data \citep{sun2020scalability,caesar2020nuscenes,wilson2023argoverse,mao2021one,yu2020bdd100k,caesar2021nuplan} and targeted evaluations for long-tail corner cases \citep{li2022coda,xu2025wod,tian2024tokenize,xie2025vlms}, but their rigid protocols prevent continuous ingestion of open-ended, real-world anomalies from heterogeneous sources (e.g., dashcams or web videos). OpenLongTail directly addresses this data-conversion bottleneck, transforming unstructured observations into synchronized, surround-view assets for robust downstream policy learning.

\paragraph{Controllable Video Generation.}

Recent world foundation models offer strong generative priors for coherent visual synthesis \citep{ali2025world,wan2025wan,yang2024cogvideox,chen2025skyreels}. Driving-oriented methods leverage these for multi-view, long-horizon, and closed-loop simulation \citep{swerdlow2024street,yang2023bevcontrol,li2024drivingdiffusion,ren2025cosmos,gao2023magicdrive,wen2024panacea,lu2024wovogen,gao2024vista,wang2025longdwm,hu2024drivingworld,gao2025magicdrive,fu2024drivegenvlm,guo2024infinitydrive,li2026far}, yet heavily depend on dense structured conditions like 3D layouts or occupancy \citep{yang2025geniedrive,lu2025infinicube}. Meanwhile, camera-controlled generation can synthesize novel viewpoints from sparse observations \citep{yu2025trajectorycrafter,bai2025recammaster,yang2026neoverse,chen2024unimlvg,lu2024seeing,yao2024mygo,guo2025dist}, but struggles in autonomous driving contexts where surround-view rig cameras exhibit minimal overlap. This sparse overlap leaves massive target regions entirely unobservable and unrecoverable via standard geometric projection or warping. To resolve this, we introduce OpenLongTail. By coupling pose-aware geometry recovery with controllable world generation, OpenLongTail hallucinates unobservable rig views while strictly enforcing cross-view geometric alignment and temporal coherence.

%% file: sections/05_methodology.tex
\section{Methodology}
\label{sec:methodology}
\begin{figure*}[t]
    \centering
    \includegraphics[width=\textwidth]{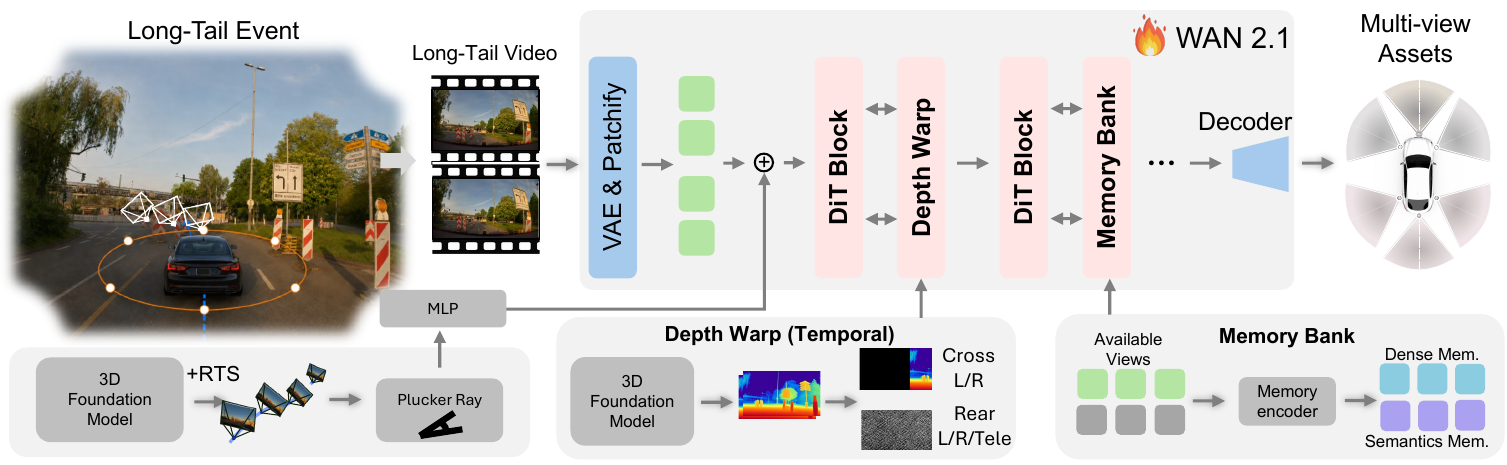}
    \caption{
    \textbf{Overview of OpenLongTail.}
    Given a long-tail front-view driving video, OpenLongTail first recovers a motion-consistent metric ego-trajectory and uses it to construct pose-aware geometric conditions. 
    The generation model combines Plücker-ray geometry, temporal depth warping, and a cross-view memory bank within a Wan 2.1-VACE backbone to synthesize synchronized non-front views under the target camera rig.
    The resulting surround-view rollout provides geometry-grounded multi-view assets for scaling downstream driving policy training.
    }
    \vspace{-2mm}
    \label{fig:pipeline}
\end{figure*}

 In this section, we present OpenLongTail, a pose-aware extrapolative-view synthesis pipeline that transforms a monocular, long-tail front-view driving video into a synchronized multi-view videos under a target camera rig (Figure~\ref{fig:pipeline}). Our approach is structured around four primary stages. First, we formulate the task as pose-informed multi-view generation conditioned on a continuous ego-trajectory. Second, we recover and stabilize a metric-scale camera trajectory from unconstrained video inputs to establish absolute spatial anchoring. Third, we execute multi-view extrapolative synthesis by anchoring a generative backbone with unified 3D ray geometries, temporal lookback depth-warps, and a topological cross-view graph memory. Finally, we optimize this entire geometry-grounded pipeline using a flow-matching training objective.

\subsection{Problem Statement}
Our objective is to transform heterogeneous long-tail monocular driving videos 
$\mathbf{x}^{\mathrm{mono}}_{1:T}$ into motion-consistent multi-view assets 
$\hat{\mathbf{X}}^{\mathcal{R}}_{1:T}$ under a target camera rig $\mathcal{R}$, 
thereby enabling scalable long-tail data construction for driving policies. 
We formulate this transformation as a two-stage generative process factorized over the continuous ego-trajectory $\mathbf{p}_{1:T}$:
\begin{equation}
P\!\left(
\hat{\mathbf{X}}^{\mathcal{R}}_{1:T}
\mid
\mathbf{x}^{\mathrm{mono}}_{1:T}, \mathcal{R}
\right)
=
\underbrace{
P\!\left(
\mathbf{p}_{1:T}
\mid
\mathbf{x}^{\mathrm{mono}}_{1:T}
\right)
}_{\substack{\text{Camera Trajectory}\\\text{Recovery}}}
\;\cdot\;
\underbrace{
P_{\theta}\!\left(
\hat{\mathbf{X}}^{\mathcal{R}}_{1:T}
\mid
\mathbf{x}^{\mathrm{mono}}_{1:T},
\mathbf{p}_{1:T},
\mathcal{R}
\right)
}_{\substack{\text{Geometry-Grounded}\\\text{Generation}}}.
\end{equation}
We instantiate the first term by recovering metric camera poses from the input monocular video using MapAnything \citep{keetha2025mapanything}, followed by forward-backward motion smoothing to suppress frame-level pose jitter while preserving metric scale. 
The resulting $\mathbf{p}^{*}_{1:T}$ provides stable pose conditioning for the second term, where we synthesize extrapolative views under the target rig using geometry priors:
\begin{equation}
\hat{\mathbf{X}}^{\mathcal{R}}_{1:T}
\sim
P_{\theta}\!\left(
\cdot
\mid
\mathbf{x}^{\mathrm{mono}}_{1:T},
\mathbf{p}^*_{1:T},
\mathcal{R};
\mathbf{W}_{1:T},
\mathbf{G}^{\mathrm{pl}},
\mathbf{M}_{1:T}
\right),
\end{equation}
where $\mathbf{W}_{1:T}$ denotes temporal depth warps that provide geometry-aligned visual evidence, 
$\mathbf{G}^{\mathrm{pl}}$ denotes Pl\"ucker ray conditioning that encodes target camera geometry, 
and $\mathbf{M}_{1:T}$ denotes cross-view memory for propagating information across generated views. 
Together, these components convert long-tail videos into coherent, synchronized multi-view assets for downstream policy learning.

\subsection{Metric-Scale Ego-Trajectory Recovery}
To construct pose-aligned multi-view assets, we extract a metric-scale camera trajectory from an input video $\mathcal{V}_{1:T}$ using MapAnything:
\begin{equation}
\label{eq:mapanything-trajectory}
\hat{\mathcal{T}}_{1:T} = \Phi_{\mathrm{MA}}(\mathcal{V}_{1:T}) = \{\hat{\mathbf{T}}_t\}_{t=1}^{T}
\end{equation}
where each pose $\hat{\mathbf{T}}_t \in SE(3)$. Because frame-level poses often contain high-frequency fluctuations that destabilize view synthesis, we refine them through a motion-consistent stabilization stage:
\begin{equation}
\label{eq:smoothed-mapanything-trajectory}
\tilde{\mathcal{T}}_{1:T}
=
\Psi_{\mathrm{RTS}}\!\left(
    \Psi_{\mathrm{KF}}\!\left(
        \hat{\mathcal{T}}_{1:T}
    \right)
\right).
\end{equation}
Specifically, a forward Kalman filter ($\Psi_{\mathrm{KF}}$) suppresses local noise, while a backward Rauch-Tung-Striebel smoother ($\Psi_{\mathrm{RTS}}$) uses the complete clip to stabilize the global path. The output trajectory $\tilde{\mathcal{T}}_{1:T}$ preserves metric-scale ego-motion while reducing jitter, yielding smoother pose conditions.

\subsection{Pose-Informed Extrapolative Synthesis}
We synthesize five non-front views $\{\mathbf{z}_t\}_{t=1}^5$ from a front-view latent $\mathbf{z}_0$, camera calibration intrinsics and extrinsics $(K, E)$, an ego-trajectory $T_{\text{anchor}}$, and a text prompt $\mathbf{c}_{\text{txt}}$. Generation is performed in the latent space of a frozen Wan 2.1 VAE, using a Wan 2.1-VACE DiT 1.3B backbone \citep{jiang2025vace}. To ground generation in the physical structure of the driving scene, we integrate three core conditioning modules into the backbone: a Geometry Encoder, a Temporal Depth Warp, and a Cross-View Memory Bank. All trainable parameters are limited to LoRA \cite{hu2022lora} adapters on the Wan-DiT/VACE self-attention layers and the three conditioning modules.

\paragraph{Geometry Conditioning.}
The Geometry Encoder injects per-token camera-ray geometry. For every spatio-temporal token at latent grid position $(h,w)$ on view $v$ at frame $\tau$, we compute its 3D ray in the ego-anchor frame as a Plücker ray:
\begin{equation}
\mathbf{r}^{(v,\tau)}_{h,w} = \bigl(\mathbf{d}^{(v,\tau)}_{h,w}; \boldsymbol{\mu}^{(v,\tau)}_{h,w}\bigr) \in \mathbb{R}^6,\qquad \boldsymbol{\mu} = \mathbf{o} \times \mathbf{d},
\end{equation}
where $\mathbf{d}$ is the unit ray direction and $\boldsymbol{\mu}$ is its moment. Together with discrete embeddings for camera identity, stream role (target or condition), and an ego-motion MLP, the Geometry Encoder produces a unified token-wise bias:
\begin{equation}
\mathbf{g}^{(v,\tau)}_{h,w} = \phi_{\text{Plücker}}\bigl(\mathbf{r}^{(v,\tau)}_{h,w}\bigr) + \mathbf{e}_{\text{cam}}(v) + \mathbf{e}_{\text{role}}(r) + \phi_{\text{traj}}\bigl(T_{\text{anchor}}\bigr) \in \mathbb{R}^{d}.
\end{equation}
Crucially, $\mathbf{g}$ is broadcast-added at three parallel entry points: the main DiT hidden state, the VACE control-branch hidden state, and the memory condition bank. This shared encoding ensures that all network branches operate in a consistent 3D ray frame.

\paragraph{Temporal Depth Warp.}
To provide pixelwise RGB priors aligned to each target viewpoint, the depth warping module constructs a geometrically aligned prior $(\mathbf{z}^{\text{warp}}_t, \mathbf{m}_t)$ for each target view $t$. This prior comprises frontal pixels reprojected into the target frame $t$ (encoded via the Wan VAE) alongside a tokenwise visibility mask. The module contains no trainable parameters, relying entirely on depth extracted from a frozen DepthCrafter and analytic warping operations. The specific reprojection strategy depends on the spatial overlap between cameras. For lateral cameras ($t \in \{1,2\}$), which share overlapping fields of view with the frontal camera, we apply a spatial warp within the same frame. A frontal pixel $\mathbf{u}$ at frame $\tau$ is unprojected using its depth $d_0$ and transformed via the relative $\text{SE}(3)$ pose:
\begin{equation}
\Pi^{\text{same}}_{0 \to t}(\mathbf{u},\tau) = K_t (E_t^{-1}E_0) K_0^{-1} \mathbf{u} d_0(\mathbf{u},\tau).
\end{equation}
In contrast, the rear cameras ($t \in \{3,4,5\}$) share no direct overlap with the frontal view. To resolve this, we exploit the forward egomotion of the vehicle by applying a temporal offset warp, sampling from a past frontal frame that spatially covers the current target $t$:
\begin{equation}
\Pi^{\text{look}}_{0 \to t}(\mathbf{u},\tau;\Delta_t) = K_t \bigl(T^{(t,\tau)}_{\text{cam}}\bigr)^{-1} T^{(0,\tau-\Delta_t)}_{\text{cam}} K_0^{-1} \mathbf{u} d_0(\mathbf{u},\tau-\Delta_t).
\end{equation}
Instead of a fixed temporal window, we utilize frame offsets $\Delta_t$ specific to each view. This design enables cameras with narrower fields of view to draw from further back along the trajectory, effectively maximizing valid spatial coverage across all targets.
\paragraph{Cross-View Memory Bank.}
While the depth warp provides pose-aligned content, maintaining seam level consistency across adjacent cameras requires each target view to condition on previously generated spatial neighbors. We encode this structure with a directed autoregressive dependency graph over the five target cameras: $\mathcal{G}(1)=\mathcal{G}(2)=\{0\}$, $\mathcal{G}(3)=\{0,1\}$, $\mathcal{G}(4)=\{0,2\}$, and $\mathcal{G}(5)=\{0,3,4\}$, where $\mathcal{G}(i)$ denotes the conditioning views used to synthesize target view $i$. Inference traverses $\mathcal{G}$ in topological order, thereby propagating consistent geometry across views.

At every forward pass, up to three conditioning latents in $\mathcal{G}(t)$ are patchified, combined with the Geometry Encoder bias, and processed through $N$ adapter blocks to produce a dense memory $\mathbf{M}^{\text{dense}} \in \mathbb{R}^{3 \times T \times H'W' \times d}$. A 64-query Perceiver-style resampler compresses this representation into a semantic memory $\mathbf{M}^{\text{sem}} \in \mathbb{R}^{3 \times T \times 64 \times d}$. At a sparse set of selected DiT layers, the target-stream tokens $\mathbf{h}^{\mathrm{tgt}}_{\tau}$ cross-attend to a short temporal window of both memories:
\begin{equation}
{\small
\Delta_{\tau}
=
g_{\mathrm{d}}(\sigma)\,
\Phi^{\mathrm{dense}}\!\left(
\mathbf{h}^{\mathrm{tgt}}_{\tau},
\mathbf{M}^{\mathrm{dense}}_{\mathcal{N}(\tau)}
\right)
+
g_{\mathrm{s}}(\sigma)\,
\Phi^{\mathrm{sem}}\!\left(
\mathbf{h}^{\mathrm{tgt}}_{\tau},
\mathbf{M}^{\mathrm{sem}}_{\mathcal{N}(\tau)}
\right),
\;
\mathbf{h}^{\mathrm{tgt}}_{\tau}
\leftarrow
\mathbf{h}^{\mathrm{tgt}}_{\tau}
+
\Delta_{\tau}.
}
\end{equation}
where $\mathcal{N}(\tau)$ denotes a local temporal neighborhood around frame $\tau$. The dense pathway anchors pixel-level correspondence across views, while the semantic pathway provides a content-level summary that remains robust to spatial misalignment.

\subsection{Training Objective} \label{sec:training}
We optimize the target stream using a flow-matching objective. For a sampled target camera $v \in \{1,\ldots,5\}$ with clean video latent $\mathbf{z}_v$, we define the noisy state $\mathbf{z}^{(\sigma)}_v = (1-\sigma)\mathbf{z}_v + \sigma\boldsymbol{\epsilon}$ and target velocity $\mathbf{v}_v = \boldsymbol{\epsilon} - \mathbf{z}_v$, given noise level $\sigma \sim p(\sigma)$. The model $f_\theta$ predicts the velocity $\hat{\mathbf{v}}_v$ conditioned on $\mathbf{z}^{(\sigma)}_v$, $\sigma$, the observed front-view latent $\mathbf{z}_0$, text $\mathbf{c}_{\mathrm{txt}}$, and view-specific geometry and control features ($\mathbf{c}^{\mathrm{VACE}}_v$, $\mathbf{M}^{\mathrm{dense}}_v$, $\mathbf{M}^{\mathrm{sem}}_v$, $\mathbf{g}_v$). The training loss minimizes the mean-squared error:
\begin{equation}
\mathcal{L}(\theta) = \mathbb{E}_{v,\sigma,\boldsymbol{\epsilon}} \left[ w_v \left\| \hat{\mathbf{v}}_v - \mathbf{v}_v \right\|_2^2 \right],
\end{equation}
where $w_v=1$, except for the rear-tele view which is weighted by $w_5=\lambda_{\mathrm{rear}}$.

%% file: sections/06_evaluation.tex
\section{Evaluation}
\label{sec:evaluation}
We evaluate OpenLongTail as a generative data engine for scaling VLA policies with heterogeneous long-tail driving videos. Our evaluation is organized around the end task first: whether synthesized long-tail multi-view data improves closed-loop driving behavior. We then analyze the intermediate components that make this improvement possible, including pose-informed extrapolative view synthesis and metric-scale ego-trajectory recovery.
Specifically, we first evaluate closed-loop driving performance in AlpaSim~\citep{alpasim_2025}, then assess generation quality under seen-scenario, unseen-scenario, and Waymo E2E~\citep{xu2025wod} splits using visual and cross-view consistency metrics, and finally measure pose accuracy and temporal coherence against recent trajectory estimation methods.

\newcommand{\nometric}{\cellcolor{gray!20}\textcolor{gray!60}{}}

\begin{table*}[t]
\centering
\scriptsize
\caption{\textbf{Main result: Closed-loop AlpaSim evaluation in long-tail events.} We report AlpaSim Score (AS, higher is better) and Collision Rate (CR, lower is better) across four representative long-tail categories and the overall average. SFT denotes supervised fine-tuning; 10K Rand denotes 10K randomly sampled training trajectories; NV-OOD denotes the NVIDIA ood subset; GT and Syn denote ground-truth and OpenLongTail synthesized multi-view data, respectively. Complex Int. and Uncommon Veh. denote complex intersections and uncommon vehicles. The best and second-best results are shown in \textbf{bold} and \underline{underlined}, respectively.}
\label{tab:closed_loop_longtail_crossview}
\setlength{\tabcolsep}{3.5pt}
\renewcommand{\arraystretch}{0.95}

\begin{adjustbox}{max width=\textwidth}
\begin{tabular}{@{}l c cc cc cc cc cc cc cc@{}}
\toprule
\multirow[c]{3}{*}{\textbf{Model}}
& \multicolumn{5}{c}{\textbf{Additional Training Data}}
& \multicolumn{8}{c}{\textbf{Representative Long-tail Events}}
& \multicolumn{2}{c}{\textbf{Avg.}} \\
\cmidrule(lr){2-6}
\cmidrule(lr){7-14}
\cmidrule(lr){15-16}
&
\multirow[c]{2}{*}{\makecell[c]{\textbf{10K}\\\textbf{Rand}}}
& \multicolumn{2}{c}{\textbf{NV-OOD}}
& \multicolumn{2}{c}{\textbf{Waymo-E2E}}
& \multicolumn{2}{c}{\textbf{Complex Int.}}
& \multicolumn{2}{c}{\textbf{Cyclists}}
& \multicolumn{2}{c}{\textbf{Uncommon Veh.}}
& \multicolumn{2}{c}{\textbf{Work Zone}}
& \multicolumn{2}{c}{} \\
\cmidrule(lr){3-4}
\cmidrule(lr){5-6}
\cmidrule(lr){7-8}
\cmidrule(lr){9-10}
\cmidrule(lr){11-12}
\cmidrule(lr){13-14}
&
& \textbf{GT}
& \textbf{Syn}
& \textbf{GT}
& \textbf{Syn}
& \textbf{AS}$\uparrow$ & \textbf{CR}$\downarrow$
& \textbf{AS}$\uparrow$ & \textbf{CR}$\downarrow$
& \textbf{AS}$\uparrow$ & \textbf{CR}$\downarrow$
& \textbf{AS}$\uparrow$ & \textbf{CR}$\downarrow$
& \textbf{AS}$\uparrow$ & \textbf{CR}$\downarrow$ \\
\midrule

Alpamayo R1~\citep{wang2025alpamayo}
&  &  &  &  &
& 0.457 & 71.4\%
& 0.534 & \underline{50.0\%}
& 0.575 & 50.0\%
& 0.683 & \underline{50.0\%}
& 0.534 & 58.8\% \\

Alpamayo 1.5~\citep{wang2025alpamayo}
&  &  &  &  &
& 0.469 & 75.0\%
& 0.347 & 100.0\%
& 0.517 & \underline{33.3\%}
& 0.731 & \underline{50.0\%}
& 0.501 & 61.1\% \\

\midrule

\multirow[c]{5}{*}{\makecell[c]{Alpamayo R1\\+ SFT\\\citep{wang2025alpamayo}}}
& \cmark &  &  &  &
& 0.659 & 25.0\%
& 0.670 & \textbf{0.0\%}
& 0.733 & \textbf{0.0\%}
& 0.936 & \textbf{0.0\%}
& 0.716 & 11.1\% \\

& \cmark & \cmark &  &  &
& \underline{0.743} & \textbf{0.0\%}
& 0.688 & \textbf{0.0\%}
& \underline{0.758} & \textbf{0.0\%}
& 0.938 & \textbf{0.0\%}
& \textbf{0.764} & \textbf{0.0\%} \\

& \cmark &  & \cmark &  &
& \textbf{0.747} & \textbf{0.0\%}
& 0.662 & \textbf{0.0\%}
& 0.717 & \textbf{0.0\%}
& 0.935 & \textbf{0.0\%}
& \underline{0.748} & \textbf{0.0\%} \\

\addlinespace[1pt]
\cmidrule(lr){2-16}
\addlinespace[1pt]

& \cmark & \cmark &  & \cmark &
& 0.691 & \underline{12.5\%}
& \textbf{0.691} & \textbf{0.0\%}
& 0.735 & \textbf{0.0\%}
& \textbf{0.958} & \textbf{0.0\%}
& 0.736 & \underline{5.6\%} \\

& \cmark &  & \cmark &  & \cmark
& 0.699 & \underline{12.5\%}
& \underline{0.689} & \textbf{0.0\%}
& \textbf{0.765} & \textbf{0.0\%}
& \underline{0.950} & \textbf{0.0\%}
& \underline{0.748} & \underline{5.6\%} \\

\bottomrule
\end{tabular}
\end{adjustbox}
\vspace{-2mm}
\end{table*}

\begin{figure}[t]
    \centering
    \includegraphics[width=\linewidth]{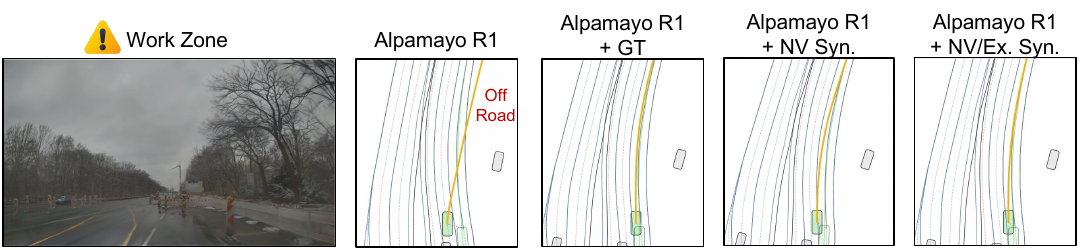}
    \caption{\textbf{Qualitative closed-loop evaluation in a work-zone long-tail event.}
Left: front-view observation of a rare work-zone scene. Right: BEV rollouts of Alpamayo R1 under different long-tail training data, where NV denotes NVIDIA LT data and Ex. denotes Waymo E2E data.}
    \label{fig:qualitative_comparison_alpasim}
    \vspace{-2mm}
\end{figure}

\subsection{Experimental Setup and Implementation}

We curate large-scale driving video logs consisting of over 200K clips from about 50K scenes across NVIDIA PhysicalAI-Autonomous-Vehicles~\citep{nvidia_physicalai_av_2025}, PandaSet~\citep{xiao2021pandaset}, and nuScenes~\citep{caesar2020nuscenes}. 
We finetune Wan2.1-VACE-1.3B as the video generation backbone using 32 NVIDIA H200 GPUs for approximately 96 hours. 
Refer to supplementary for the implementation of finetuning and evaluation. We will open-source the code and release the model checkpoints.

\subsection{Downstream Closed-Loop Policy Performance}

We first evaluate whether OpenLongTail-generated assets improve downstream closed-loop driving behavior.
For downstream SFT and AlpaSim \citep{alpasim_2025} evaluation, we follow the Alpamayo-R1 \citep{wang2025alpamayo} camera interface, which consumes four views as input.
Specifically, the front view is taken from the original ground-truth video, the front-tele view is obtained by applying a deterministic FoV-Z crop to the front view with a $30^\circ$ field of view, and the cross-left and cross-right views are synthesized by our pose-informed generation model under the target camera rig.
The ego-motion used for both generation conditioning and policy input is recovered by our ego-trajectory module.

To evaluate the downstream performance of the fine-tuned VLA policies, we conduct closed-loop simulations on 53 long-tail events in AlpaSim~\citep{alpasim_2025}. 
At each simulation step, AlpaSim renders the corresponding multi-camera observations through its 3D reconstruction and novel-view-synthesis pipeline according to the current ego state, and feeds the resulting observations back to the VLA policy for action prediction. 
This closed-loop setting deploys the policy as an ego-driver whose continuous actions update the simulator state, directly testing robustness to covariate shift and compounding errors in long-tail events. 
For each policy and event, we run two independent rollouts and report the average metrics. 
As shown in Table~\ref{tab:closed_loop_longtail_crossview}, the baselines struggle in these long-tail scenarios, yielding low AS and high CR. 
In contrast, fine-tuning with NVIDIA synthesized assets (NV Syn) raises the average AS from 0.534 to 0.748 at a 0.0\% collision rate, comparable to ground-truth multi-view data (NV GT, 0.764), showing that OpenLongTail-generated assets are a effective alternative to multi-camera capture for long-tail policy training.

\begin{table}[t]
\centering
\scriptsize
\caption{\textbf{Unseen-scene extrapolative view evaluation.}
We evaluate target-view synthesis on held-out unseen driving scene.
We compare OpenLongTail with novel trajectory video synthesis baselines and report PSNR and LPIPS for each target view, together with GeoKPM $\mu$ for cross-view geometric consistency, where values closer to 100 indicate better consistency.}
\label{tab:wm_ood}
\setlength{\tabcolsep}{2.0pt}
\renewcommand{\arraystretch}{0.92}
\begin{adjustbox}{width=\columnwidth}
\begin{tabular}{@{}l*{5}{cc}c@{}}
\toprule
\multirow[c]{2}{*}{\textbf{Method}}
& \multicolumn{2}{c}{\textbf{Cross-left}}
& \multicolumn{2}{c}{\textbf{Cross-right}}
& \multicolumn{2}{c}{\textbf{Rear-left}}
& \multicolumn{2}{c}{\textbf{Rear-right}}
& \multicolumn{2}{c}{\textbf{Rear-tele}}
& \multirow[c]{2}{*}{\textbf{GeoKPM $\mu$}} \\
\cmidrule(lr){2-3}
\cmidrule(lr){4-5}
\cmidrule(lr){6-7}
\cmidrule(lr){8-9}
\cmidrule(lr){10-11}
& PSNR $\uparrow$ & LPIPS $\downarrow$
& PSNR $\uparrow$ & LPIPS $\downarrow$
& PSNR $\uparrow$ & LPIPS $\downarrow$
& PSNR $\uparrow$ & LPIPS $\downarrow$
& PSNR $\uparrow$ & LPIPS $\downarrow$
& \\
\midrule
TrajectoryCrafter~\citep{yu2025trajectorycrafter}
& 10.02 & 0.734
& 9.70  & 0.725
& 9.79  & 0.736
& 9.14  & 0.751
& 9.29  & 0.768
& 10.77 \\

Gen3C~\citep{ren2025gen3c}
& 12.08 & 0.649
& 11.74 & 0.675
& 11.24 & 0.738
& 10.99 & 0.751
& 11.52 & 0.750
& 8.61 \\

ReCamMaster~\citep{bai2025recammaster}
& 12.20 & 0.677
& 11.91 & 0.703
& 11.12 & 0.754
& 10.98 & 0.760
& 11.36 & 0.756
& 16.58 \\

Vista4D~\citep{lin2026vista4d}
& 11.00 & 0.677
& 11.05 & 0.686
& 10.73 & 0.725
& 10.03 & 0.745
& 10.77 & 0.737
& 18.86 \\
\midrule
\textbf{Ours}
& \textbf{13.28} & \textbf{0.597}
& \textbf{12.38} & \textbf{0.628}
& \textbf{12.21} & \textbf{0.643}
& \textbf{11.71} & \textbf{0.643}
& \textbf{11.80} & \textbf{0.639}
& \textbf{82.41} \\
\bottomrule
\end{tabular}
\end{adjustbox}
\vspace{-4mm}
\end{table}

\begin{figure}[t]
    \centering
    \includegraphics[width=\linewidth]{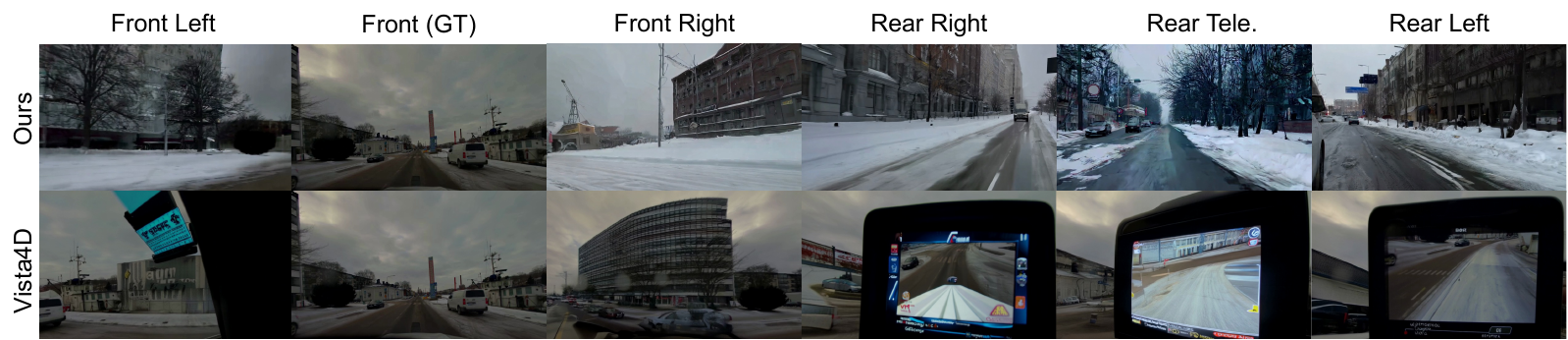}
    \caption{\textbf{Qualitative comparison of generated multi-view driving videos.}
    We compare OpenLongTail with Vista4D~\citep{lin2026vista4d}. Our model produces synchronized target views conditioned on the front camera and target camera geometry, preserving scene layout and cross-view consistency across side and rear cameras. Please refer to the supplementary material for more results.}
    \label{fig:qualitative_comparison}
    \vspace{-3mm}
\end{figure}
\paragraph{Scaling to external sources.}
By converting monocular videos into target-rig assets, OpenLongTail can further unlock external data. Augmenting NV Syn. with assets synthesized from Waymo E2E~\citep{xu2025wod} front cameras (NV + Waymo Syn) complements the NV-only strategy and recovers several failure cases, improving uncommon vehicles ($0.717{\to}0.765$), cyclists ($0.662{\to}0.689$), and work zones ($0.935{\to}0.950$). The benefit is alignment-conditioned: the complex-intersection slice regresses, as Waymo under-represents the human-guided and cone-delineated cases dominant in NV (detailed in the supplementary). We further provide qualitative closed-loop rollouts in Figure~\ref{fig:qualitative_comparison_alpasim}.
In work zone event, the Alpamayo R1 baseline deviates from the drivable lane and drives off-road, while policies fine-tuned with either ground-truth multi-view data or OpenLongTail-synthesized views produce more stable trajectories aligned with the lane structure.
This qualitative result aligns with the quantitative trends in Table~\ref{tab:closed_loop_longtail_crossview}, showing that OpenLongTail improves closed-loop performance while yielding safer behavior in challenging long-tail scenarios. Please refer to supplementary for comprehensive results and analysis.

\begin{figure}[t]
    \centering
    \includegraphics[width=\linewidth]{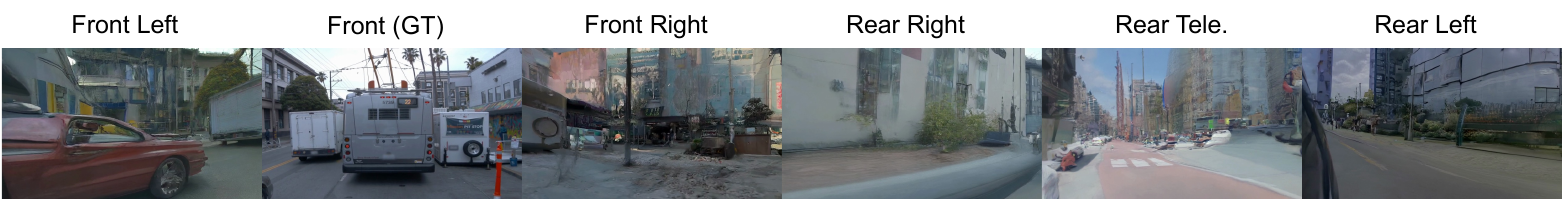}
    \caption{\textbf{Qualitative results on Waymo E2E long-tail data.}
    We apply OpenLongTail to external long-tail driving videos and synthesize synchronized target views. Given the observed front view, our model generates spatially coherent side and rear views while preserving the overall scene layout and driving context. Please refer to the supplementary material for more results.}
    \label{fig:qualitative_comparison_external_front}
    \vspace{-3mm}
\end{figure}

\begin{figure}[t]
    \centering
    \includegraphics[width=\linewidth]{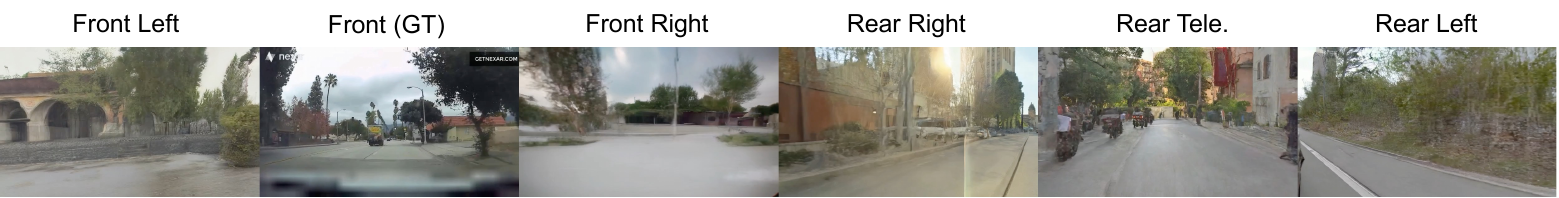}
    \caption{\textbf{Qualitative results on Nexar in-the-wild crowd-sourced dashcams.}
    We apply OpenLongTail to consumer dashcam videos. Given only the observed monocular front view, our model recovers the ego-trajectory and synthesizes synchronized extrapolative views under the target rig, preserving the overall scene layout and driving context.}
    \label{fig:qualitative_comparison_nexar}
    \vspace{-3mm}
\end{figure}
\begin{table*}[t]
\centering
\scriptsize
\setlength{\tabcolsep}{2.8pt}
\renewcommand{\arraystretch}{1.12}
\caption{
\textbf{Pose evaluation under metric-scale and Sim(3)-aligned protocols.}
Metric-scale evaluation preserves absolute scale; Sim(3) alignment removes global pose and scale before error computation.
\protect\colorbox{gray!20}{\strut Light-gray cells} denote methods that are not metric-scale.
Lower is better for ATE, RPE-r, RPE-t, RotErr, RotAnc, TrErr, Jerk, and Acc.Var; higher is better for LDJ.
}
\resizebox{\textwidth}{!}{
\begin{tabular}{l|ccccccc|cccccc}
\toprule
\multirow{2}{*}{Method}
& \multicolumn{7}{c|}{Metric-scale Evaluation}
& \multicolumn{6}{c}{Sim(3)-aligned Evaluation} \\
\cmidrule(lr){2-8} \cmidrule(lr){9-14}
& ATE$\downarrow$
& RPE-r$\downarrow$
& RPE-t$\downarrow$
& RotErr$\downarrow$
& Jerk$\downarrow$
& Acc.Var$\downarrow$
& LDJ$\uparrow$
& ATE$\downarrow$
& RotAnc$\downarrow$
& RotErr$\downarrow$
& TrErr$\downarrow$
& Jerk$\downarrow$
& Acc.Var$\downarrow$ \\
\midrule

DROID-W \citep{li2026droid}
& \nometric & \nometric & \nometric & \nometric & \nometric & \nometric & \nometric
& 0.284 & 0.861 & 0.217 & 12.99 & 4570.6 & 119.60 \\

MegaSAM \citep{li2025megasam}
& \nometric & \nometric & \nometric & \nometric & \nometric & \nometric & \nometric
& 0.444 & 0.456 & \textbf{0.129} & 15.57 & 3704.4 & 80.69 \\

VGGT \citep{wang2025vggt}
& \nometric & \nometric & \nometric & \nometric & \nometric & \nometric & \nometric
& 0.202 & 0.393 & 0.134 & 13.54 & 2947.7 & 61.19 \\

ViPE \citep{huang2025vipe}
& 3.831 & 0.137 & 59.47 & 0.472 & 929.5 & 15.51 & -9.18
& \textbf{0.162} & 0.472 & 0.137 & \textbf{9.86} & 1055.5 & 18.44 \\

MapAnything \citep{keetha2025mapanything}
& \textbf{2.212} & 0.155 & 23.52 & 0.352 & 4737.4 & 53.22 & -12.65
& 0.171 & 0.352 & 0.155 & 12.57 & 4243.9 & 47.52 \\

Ours
& \textbf{2.212} & \textbf{0.129} & \textbf{23.05} & \textbf{0.340} & \textbf{283.9} & \textbf{5.96} & \textbf{-7.48}
& \textbf{0.162} & \textbf{0.340} & \textbf{0.129} & \underline{11.24} & \textbf{254.5} & \textbf{5.33} \\

\bottomrule
\end{tabular}
}
\label{tab:pose_metric_sim3_combined}
\vspace{-4mm}
\end{table*}

\subsection{Pose-informed Extrapolative View Evaluation}

With the downstream benefits established, we next evaluate the generated multi-view videos.
We evaluate the pose-informed diffusion under three settings: seen-scenario, unseen-scenario, and external monocular-source evaluation. 
The seen-scenario split contains 20 clips from held-out temporal segments of scenes observed during training, measuring whether the model can complete missing views in familiar environments. Please refer to the supplementary for this evaluation.
The unseen-scenario split contains 218 clips from entirely unseen scenes, evaluating generalization to novel driving scenarios from the same data source. 
Finally, the external monocular split uses only front-camera videos from Waymo E2E \citep{xu2025wod}, which are never seen during training.

We report PSNR, SSIM~\citep{wang2004image}, LPIPS~\citep{zhang2018unreasonable}, FID~\citep{heusel2017gans}, and FVD~\citep{unterthiner2018towards} as evaluation metrics. 
Following Drive-WM~\citep{wang2024driving}, we evaluate cross-view consistency with keypoint matching (KPM) under epipolar-geometry filtering, avoiding shortcut solutions such as view copying.
Table~\ref{tab:wm_ood} shows that it outperforms the baselines on unseen scenes, especially in GeoKPM $\mu$, indicating stronger cross-view geometric consistency. The especially large gain in GeoKPM $\mu$ suggests that OpenLongTail does not merely improve per-view visual fidelity, but more importantly preserves a shared 3D scene structure across extrapolated target views.
We provide qualitative comparisons of generated surround-view videos in Figure~\ref{fig:qualitative_comparison}. 
For Waymo E2E \citep{xu2025wod} split, we report qualitative results in Figure~\ref{fig:qualitative_comparison_external_front}. 
Please refer to the supplementary material for more results.

We further apply OpenLongTail to Nexar \citep{nexar2025dashcamcollisionprediction}, a large-scale collection of crowd-sourced consumer dashcam videos. Unlike Waymo E2E \citep{xu2025wod}, which is external but still collected through a dedicated autonomous-driving effort with consistent camera hardware, Nexar more directly represents the ubiquitous dashcam setting. Its clips come from heterogeneous consumer devices with unknown and varying intrinsics, contain rolling-shutter and compression artifacts, lack a calibrated multi-camera rig, and provide no reliable metric camera pose. We use only the monocular front-view stream, and, as with all external sources, recover ego-motion entirely through our pose estimation module. As shown in Figure \ref{fig:qualitative_comparison_nexar}, given only a raw front-view video, OpenLongTail synthesizes synchronized extrapolative views under the target rig while preserving the dominant scene structure, including road direction, roadside layout, and surrounding traffic, and maintaining consistency across the generated cameras. Despite the degraded and uncalibrated input, the recovered trajectory still provides a stable geometric anchor for extrapolative synthesis, demonstrating that OpenLongTail generalizes from curated corpora to genuinely unconstrained consumer footage.


\subsection{Metric-Scale Pose Recovery Evaluation}

Finally, we evaluate the ego trajectory that provides pose conditioning for target-view synthesis. We conduct camera trajectory recovery evaluation on 218 clips from 109 scenes, comparing OpenLongTail with recent trajectory estimation methods, including DROID-W~\citep{li2026droid}, MegaSAM~\citep{li2025megasam}, VGGT~\citep{wang2025vggt}, MapAnything~\citep{keetha2025mapanything}, and ViPE~\citep{huang2025vipe}, under both metric-scale and Sim(3)-aligned protocols. As shown in Table~\ref{tab:pose_metric_sim3_combined}, our pipeline matches the best metric-scale ATE of MapAnything while reducing jerk from 4737.4 to 283.9 and acceleration variance from 53.22 to 5.96, indicating substantially smoother recovered trajectories without sacrificing absolute-scale accuracy. Under Sim(3) alignment, our method also matches the best ATE and achieves the lowest rotation error, jerk, and acceleration variance, with only a small gap to ViPE in translation error. These results show that OpenLongTail recovers trajectories that are both geometrically accurate and temporally stable, providing smooth and metrically useful pose conditions for target-rig view synthesis.

%% file: sections/07_conclusion.tex
\section{Conclusion and Limitations}
\label{sec:conclusion}
We present OpenLongTail, an open-scaling generative framework that transforms heterogeneous, unposed videos into synchronized, pose-grounded multi-view assets for scaling driving policies. Across closed-loop evaluation in AlpaSim \cite{alpasim_2025}, extrapolative view synthesis, and ego-trajectory recovery, OpenLongTail-synthesized assets consistently improve policy robustness under long-tail events, establishing ubiquitous monocular videos as a scalable training source. Beyond these gains, our analysis refines the notion of generative scaling: effective scaling is governed not by data \emph{quantity} alone but by data \emph{distribution}. Out-of-source data improves robustness and recovers failure cases, yet its contribution is alignment-conditioned, indicating that sample-efficient scaling requires aligning the source distribution with the target long-tail slice rather than merely enlarging the training pool. Remaining limitations include the inference cost of diffusion-based synthesis, residual temporal artifacts, and camera biases that persist across rig configurations. Future work will couple efficient inference and explicit camera-parameter conditioning with distribution-aware source selection, advancing generative scaling toward both higher fidelity and greater sample efficiency.

%% file: sections/08_appendix.tex
\section{Dataset Curation and Splits}
\label{app:dataset}

\subsection{Dataset Sources}
\label{app:dataset_sources}

We curate driving video logs from NVIDIA PhysicalAI-Autonomous-Vehicles (PAV) \citep{nvidia_physicalai_av_2025}, PandaSet \citep{xiao2021pandaset}, and nuScenes \citep{caesar2020nuscenes}, resulting in approximately 200K 41-frame clips from about 50K scenes. Most training clips are collected from PAV, which provides diverse large-scale driving logs and rich long-tail driving scenarios. PandaSet and nuScenes are further incorporated to improve sensor, camera, weather, scene-layout, and location diversity.
Each training sample contains 41 consecutive frames. For multi-view sources, we use the front camera as the observed conditioning view and use the remaining target-rig cameras as supervision for extrapolative view synthesis. For Waymo E2E \citep{xu2025wod}, we use OpenLongTail to synthesize the missing views under the target rigs.

\begin{table}[h]
\centering
\scriptsize
\caption{\textbf{Dataset sources and usage.}
We curate large-scale driving clips mainly from PAV and supplement them with PandaSet and nuScenes for additional camera and scene diversity. Waymo E2E is used only for external monocular-source evaluation and is never used for training.}
\label{tab:dataset_sources}
\setlength{\tabcolsep}{3.0pt}
\renewcommand{\arraystretch}{0.92}
\begin{adjustbox}{max width=\textwidth}
\begin{tabular}{@{}lccccc@{}}
\toprule
Source
& Approx. Clips Scale
& View Type
& Clip Length
& Primary Usage
& Split Rule \\
\midrule
PAV \citep{nvidia_physicalai_av_2025}
& 200,000
& multi-view
& 41 frames
& training / test / validation
& scene ID + temporal split \\
PandaSet \citep{xiao2021pandaset}
& 1236
& multi-view
& 41 frames
& training / validation
& scene ID + temporal split \\
nuScenes \citep{caesar2020nuscenes}
& 12,000
& multi-view
& 41 frames
& training / validation
& scene ID + temporal split \\
Waymo E2E \citep{xu2025wod}
& 1000
& front-view
& 41 frames
& test only
& not used in training \\
\bottomrule
\end{tabular}
\end{adjustbox}
\vspace{-2mm}
\end{table}

\subsection{Train/Test Split Protocol}
\label{app:split_protocol}

\paragraph{Seen-scenario split.}
The seen-scenario split evaluates temporal generalization within scenes observed during training. For a given source segment, we allocate disjoint 41-frame temporal windows for training and testing. For instance, frames 0–40 may be used for training, while a subsequent window spanning frames 42–82 is reserved for testing. Consequently, the test clips share the same scene ID and global environment as the training data, ensuring the model is evaluated on strictly unseen frames from familiar environments.

\paragraph{Unseen-scenario split.}
The unseen-scenario split evaluates generalization to novel scenes. We split data by scene ID, such that all clips from the held-out scene IDs are excluded from training. This prevents the model from observing the same scene layout, camera trajectory, or visual appearance during fine-tuning. We use this split for the main extrapolative view evaluation.

\paragraph{Waymo E2E split.}
Waymo E2E \citep{xu2025wod} is used as an external source evaluation set. We use its front-view videos as input and synthesize the missing target-rig views using OpenLongTail. No Waymo E2E clips are used for model training, validation, hyperparameter tuning, or checkpoint selection.

\section{Generative Scaling across Diverse Sources}
\label{app:generative_scaling_sources}

OpenLongTail is designed to scale long-tail driving data by converting heterogeneous videos into policy-compatible multi-view assets. In this setting, generative scaling should not be interpreted as simply increasing the number of samples from a fixed distribution. Instead, it expands the usable source pool by allowing videos from different datasets, camera setups, and collection protocols to be transformed into a shared target-rig format. This is important because each source naturally carries its own distributional bias, including road topology, ego-motion patterns, traffic-agent composition, temporary traffic-control cues, and camera viewpoints. Therefore, adding an external source provides complementary long-tail diversity, while its downstream contribution depends on how well its scene factors align with the target evaluation distribution.

\begin{figure}[t]
    \centering
    \includegraphics[width=\linewidth]{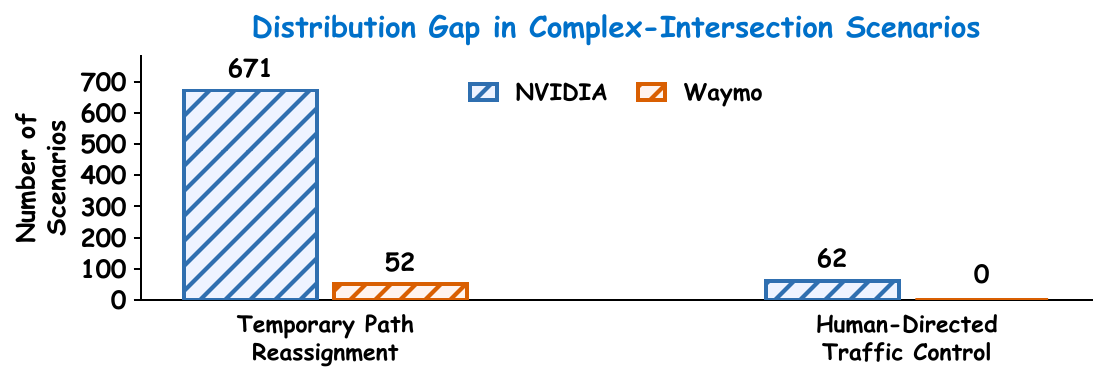}
    \caption{\textbf{Distribution gap in complex-intersection scenarios.}
    We compare the occurrence of two complex-intersection patterns in NVIDIA and Waymo data: \emph{Temporary Path Reassignment}, where cones, closures, or temporary routing cues override nominal lane markings, and \emph{Human-Guided Traffic Control}, where workers or police officers actively direct vehicle behavior. NVIDIA contains substantially higher coverage of these long-tail cases, while Waymo contains few or no examples.}
    \label{fig:distribution_gap}
    \vspace{-2mm}
\end{figure}

\begin{figure}[t]
    \centering
    \includegraphics[width=\linewidth]{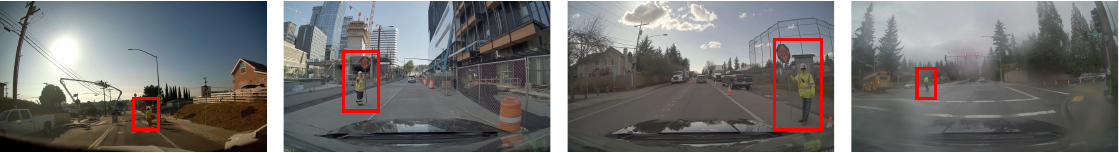}
    \caption{\textbf{Representative human-directed complex intersection examples from PAV
\citep{nvidia_physicalai_av_2025}.} Our long-tail data includes scenarios where workers or police officers actively direct traffic using signs or gestures, requiring the policy to follow temporary human guidance instead of relying only on static road geometry. These human-guided cases are largely missing from the Waymo E2E \citep{xu2025wod}.}
    \label{fig:vis_example}
    \vspace{-2mm}
\end{figure}

\begin{figure}[t]
\centering
\includegraphics[width=\linewidth]{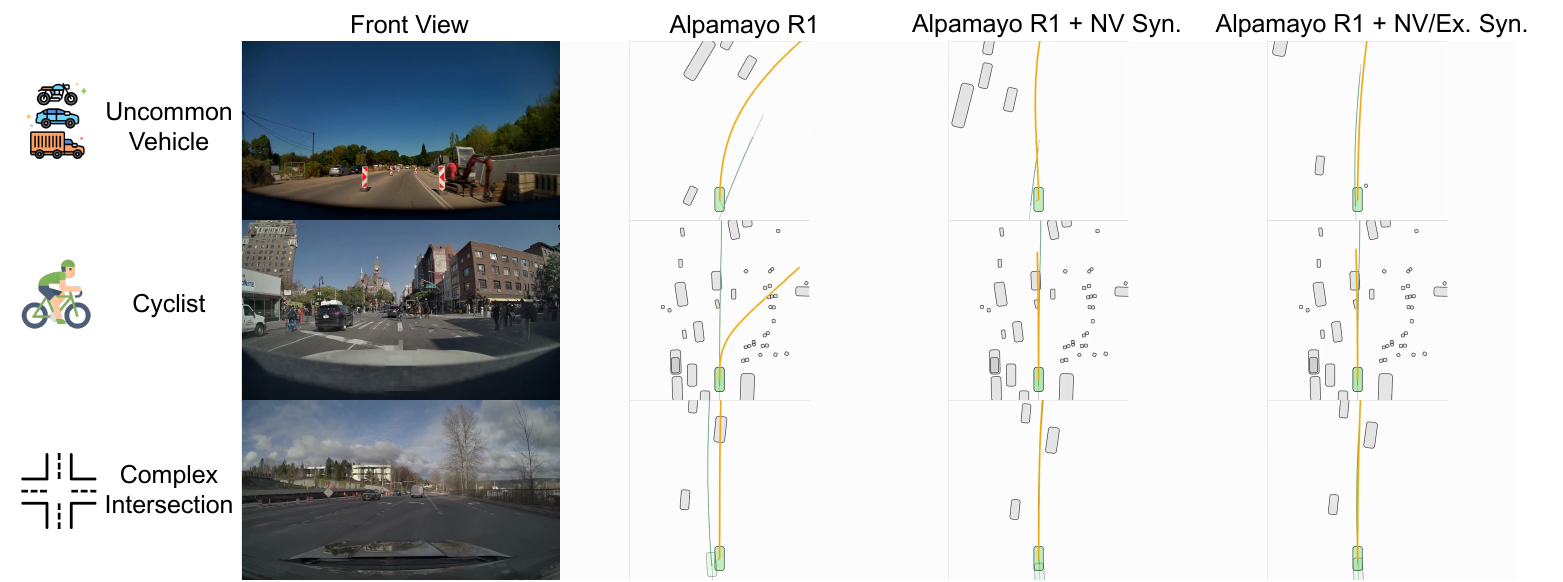}
\caption{\textbf{Closed-loop benefits of generative scaling on long-tail scenarios.}
We compare Alpamayo R1 with variants augmented by OpenLongTail-generated assets from NV data and from both NV and external sources. Across representative long-tail cases, including uncommon vehicles, cyclists, and complex intersections, adding generated assets improves the closed-loop rollout behavior. The policy trained with additional NV and external synthesized data produces trajectories that better follow the intended drivable corridor and avoid unsafe interactions. This demonstrates that generative scaling does not merely increase data volume, but expands useful long-tail coverage for downstream VLA training.}
\label{fig:closed_loop_scaling}
\vspace{-2mm}
\end{figure}

Waymo E2E \citep{xu2025wod} provides a useful external source for generative scaling. OpenLongTail converts Waymo front-camera videos into policy-compatible multi-view assets under the target camera rig, enabling these out-of-source videos to be used for downstream VLA training. Beyond qualitative conversion quality, we further evaluate whether the converted assets provide practical training benefits in closed-loop long-tail scenarios. As shown in Fig.~\ref{fig:closed_loop_scaling}, augmenting Alpamayo R1 with OpenLongTail-generated data improves rollout behavior across several representative long-tail cases, including uncommon vehicles, cyclists, and complex intersections. The model trained with both NV and external synthesized assets produces more scene-consistent trajectories than the base model, showing that heterogeneous generated data can expand useful long-tail coverage rather than simply increasing sample count.

At the same time, these gains are naturally alignment-conditioned. To better understand this effect, we analyze the complex-intersection category, where the target evaluation assets contain several fine-grained scenario factors beyond standard four-way intersections. These include curved road geometries, irregular intersection layouts, weakly marked or cone-delineated drivable paths, and human-guided traffic-control scenarios, such as police officers or construction workers directing traffic. Such cases require the policy to reason beyond static lane geometry and respond to temporary, non-standard control cues. Fig.~\ref{fig:vis_example} visualizes representative human-guided examples from PAV, where the ego vehicle must respond to temporary human instructions rather than relying only on static lane geometry. These cases represent an important class of long-tail supervision, but they are largely absent from the added Waymo source.

We quantify the source alignment between NV and Waymo by measuring the coverage of these key scenario factors. As shown in Fig.~\ref{fig:distribution_gap}, for temporary path reassignment and weakly marked complex-intersection cases, the NV subset contains 671 examples, while Waymo E2E contains 52. For human-guided traffic-control cases, the NV subset contains 62 examples, whereas Waymo E2E contains 0. These statistics show that the two sources emphasize different sub-distributions within the broad complex-intersection label. NV provides broader coverage of curved, cone-guided, and human-guided cases, while the added Waymo subset is relatively more concentrated on canonical intersection layouts with regular lane structures.

This analysis provides a more precise interpretation of generative scaling. OpenLongTail enables the use of heterogeneous videos by converting them into target-rig multi-view assets, allowing external datasets like Waymo E2E \citep{xu2025wod} to expand the long-tail training pool. However, the transfer efficiency of these added sources depends heavily on source category alignment. When an external source shares key factors with the target evaluation slice, it yields direct training benefits; when it emphasizes a different sub-distribution, it broadens overall diversity but offers limited gains for that specific slice. Ultimately, effective long-tail generative scaling requires both robust conversion capabilities and distribution-aware source selection.

\section{Implementation Details}
\label{app:implementation}

\subsection{Generation Model Fine-Tuning}
\label{app:finetuning}
We fine-tune Wan2.1-VACE-1.3B \citep{jiang2025vace} to serve as our video generation backbone. While the Wan VAE remains frozen, trainable parameters are strictly limited to LoRA adapters applied to the Wan-DiT/VACE self-attention layers, alongside our proposed conditioning modules: the geometry encoder, the temporal-depth-warp interface, and the memory bank module.

\begin{table}[t]
\centering
\scriptsize
\caption{Generation model fine-tuning configuration.}
\label{tab:finetune_config}
\setlength{\tabcolsep}{4pt}
\renewcommand{\arraystretch}{0.95}
\begin{adjustbox}{max width=\textwidth}
\begin{tabular}{@{}lc@{}}
\toprule
Item & Value \\
\midrule
\multicolumn{2}{l}{\textit{Architecture}} \\
Backbone & Wan2.1-VACE-1.3B \\
VAE & frozen Wan VAE \\
Clip length & 41 frames \\
Latent resolution & $11 \times 60 \times 104$ (T$\times$H$\times$W) \\
Target views & cross-left, cross-right, rear-left, rear-right, rear-tele \\
Trainable modules & LoRA adapters + Geometry Encoder + Memory \\
LoRA rank / alpha & 32 / 16 \\
LoRA target projections & self-attention $\{Q, K, V, O\}$ in Wan DiT and VACE blocks \\
Cross-view (Memory) layers & $\{4, 9, 14, 19, 24, 29\}$ out of 30 DiT layers \\
Semantic resampler queries & 64 \\
Graph-gate init bias & $-1.4$ \\
\midrule
\multicolumn{2}{l}{\textit{Optimization}} \\
Optimizer & AdamW (hybrid: separate groups for new modules and LoRA) \\
Learning rate (new modules) & $1\!\times\!10^{-5}$ \\
Learning rate (LoRA) & $1\!\times\!10^{-5}$ \\
LR scheduler & Cosine with warmup, min-LR ratio $0.1$ \\
Warmup steps & 500 \\
Total training steps & 20{,}000 \\
Gradient clip & $0.5$ \\
Global batch size & 32 clips (32 GPUs $\times$ 1 clip / GPU) \\
\midrule
\multicolumn{2}{l}{\textit{Compute}} \\
Training GPUs & 32 $\times$ NVIDIA H200 (4 nodes $\times$ 8 GPUs) \\
Training time & $\sim$96 hours \\

\bottomrule
\end{tabular}
\end{adjustbox}
\vspace{-2mm}
\end{table}

\paragraph{Sigma-Dependent Memory Gates.}
The dense and semantic memory branches are modulated by noise-dependent gates $g_d(\sigma)$ and $g_s(\sigma)$, where $\sigma$ denotes the flow-matching noise level. These gates control the strength of cross-view memory injection at various denoising stages, enabling the model to adaptively balance semantic context with spatial correspondence.

At each Memory layer $\ell$, the gates are generated by a lightweight, layer-specific MLP:
\begin{equation}
    [g_d(\sigma), g_s(\sigma)] = \operatorname{Sigmoid}(\psi_\ell(\sigma)), \quad \text{where} \quad \psi_\ell = \mathrm{Linear}_2 \circ \mathrm{SiLU} \circ \mathrm{Linear}_1.
\end{equation}
The MLP is evaluated on a per-sample basis, ensuring each clip receives gates conditioned on its own sampled noise level. The output gates are scalars in the range $(0, 1)$ and are applied to their corresponding dense and semantic memory updates.

We initialize the final projection of $\psi_\ell$ with zero weights and a negative bias, yielding an initial gate value of approximately $0.2$. As a result, the Memory initially acts as a weak residual pathway, preserving the pretrained Wan-VACE control branch at the onset of training. Because these gates are not shared across the six Memory layers, each layer can independently learn a distinct noise-dependent injection schedule. This design extends the stabilization principle of zero-initialized ControlNet-style residual branches by introducing diffusion-time-dependent memory modulation.

\section{Inference Pipeline}
\label{app:inference}

Algorithm~\ref{alg:openlongtail_inference} summarizes the inference procedure. Given a 41-frame front-view video, OpenLongTail first recovers an ego trajectory, estimates depth, constructs temporal depth-warp conditions, encodes target-camera Pl\"ucker rays, and then synthesizes the five non-front views in an autoregressive topological order. Generated views are added back to the cross-view memory bank and used as spatial context for subsequent target cameras.

\begin{algorithm}[t]
\caption{OpenLongTail inference from one front-view video}
\label{alg:openlongtail_inference}
\begin{algorithmic}[1]
\Require Front-view video $x^0_{1:T}$, target rig $\mathcal{R}$, camera intrinsics/extrinsics $(K,E)$, text prompt $c_{\rm txt}$, clip length $T=41$
\Ensure Synthesized multi-view rollout $\hat{X}^{\mathcal{R}}_{1:T}$

\State Recover metric ego trajectory $\tilde{T}_{1:T}$ from $x^0_{1:T}$
\State Apply Kalman filtering and RTS smoothing to stabilize $\tilde{T}_{1:T}$
\State Estimate front-view depth $D^0_{1:T}$ using the frozen depth model
\State Encode the front-view video into latent $z^0$ using the frozen Wan VAE
\State Initialize memory bank $\mathcal{M} \leftarrow \{z^0\}$
\State Define target-view dependency graph:
\Statex \[
\mathcal{G}(1)=\mathcal{G}(2)=\{0\},\quad
\mathcal{G}(3)=\{0,1\},\quad
\mathcal{G}(4)=\{0,2\},\quad
\mathcal{G}(5)=\{0,3,4\}.
\]
\For{target view $v$ in topological order $(1,2,3,4,5)$}
    \State Construct Pl\"ucker-ray geometry condition $G_{\rm pl}^{v}$ from $(K_v,E_v,\tilde{T}_{1:T})$
    \If{$v \in \{1,2\}$}
        \State Construct same-frame depth warp $W^v_{1:T}$ from front view to target view $v$
    \Else
        \State Construct temporal lookback depth warp $W^v_{1:T}$ using view-specific offset $\Delta_v$
    \EndIf
    \State Retrieve conditioning latents from $\mathcal{M}$ according to $\mathcal{G}(v)$
    \State Build dense memory $M_{\rm dense}^{v}$ and semantic memory $M_{\rm sem}^{v}$
    \State Sample target latent $\hat{z}^{v}$ with the pose-conditioned Wan2.1-VACE diffusion model
    \State Decode $\hat{z}^{v}$ into RGB video $\hat{x}^{v}_{1:T}$ using the frozen Wan VAE decoder
    \State Update memory bank $\mathcal{M} \leftarrow \mathcal{M} \cup \{\hat{z}^{v}\}$
\EndFor
\State Return $\hat{X}^{\mathcal{R}}_{1:T}=\{x^0_{1:T},\hat{x}^{1}_{1:T},\ldots,\hat{x}^{5}_{1:T}\}$
\end{algorithmic}
\end{algorithm}

\section{Evaluation Protocol}
\label{app:evaluation_protocol}

\subsection{Image and Video Fidelity Metrics}
\label{app:fidelity_metrics}

 We report PSNR, SSIM, LPIPS, FID, and FVD. PSNR, SSIM, and LPIPS are computed between the generated target-view frames and the corresponding ground-truth target-view frames after resizing all methods to the same evaluation resolution. Unless otherwise stated, these metrics are computed over the full target-camera image. FID is computed over generated and ground-truth frames aggregated across the evaluation split. FVD is computed over 41-frame generated and ground-truth clips. For the in-distribution results, we follow the seen-scenario split defined in Sec.~\ref{app:split_protocol}, where test clips are temporally held-out but share scene IDs with the training data. 

\subsection{Cross-View Geometric Consistency}
\label{app:kpm}

We evaluate cross-view consistency with GeoKPM, a geometry-aware variant of the keypoint matching metric used in Drive-WM~\citep{wang2024driving}. Given a synchronized generated view pair $(\hat I_{t,a}, \hat I_{t,b})$, we first run LoFTR~\citep{sun2021loftr} and keep confident matches above a fixed confidence threshold. Let $M(\hat I_{t,a}, \hat I_{t,b})$ denote the number of raw confident matches, and let $M_{\rm geo}(\hat I_{t,a}, \hat I_{t,b})$ denote the subset of these matches that satisfy the epipolar geometry induced by the known camera intrinsics, extrinsics, and recovered ego pose. A match is considered geometrically valid if its Sampson error under the corresponding fundamental matrix is below a fixed threshold $\tau_{\rm epi}$.

We define GeoKPM as the fraction of generated matches that are geometrically valid:
\begin{equation}
\mathrm{GeoKPM}
=
\frac{1}{|\mathcal{S}|}
\sum_{(t,a,b)\in\mathcal{S}}
\frac{
M_{\rm geo}(\hat I_{t,a}, \hat I_{t,b})
}{
M(\hat I_{t,a}, \hat I_{t,b})
}
\times 100\% .
\label{eq:geokpm}
\end{equation}

Unlike raw KPM, which only measures the abundance of LoFTR matches, GeoKPM measures whether the generated correspondences are consistent with the true cross-view camera geometry. This distinction is important because shortcut solutions such as copying the front view into a target view can produce many raw matches, but these matches generally violate the target-camera epipolar constraint. GeoKPM therefore penalizes front-copy artifacts and provides a more direct measure of geometric consistency across generated views.

In our experiments, all methods are evaluated using the same camera parameters, recovered ego pose, output resolution, LoFTR confidence threshold, and epipolar threshold. We use a LoFTR confidence threshold of $0.6$ and set $\tau_{\rm epi}=10$ pixels.
\subsection{Baseline Evaluation Fairness}
\label{app:baseline_fairness}

All baseline methods are evaluated under the same input front-view videos, target camera rig, recovered ego trajectory, output resolution, and clip length. We use the same diffusion sampling steps, which are 50 steps across different methods. If a baseline uses a fixed public inference schedule, we follow its official inference setting and resize outputs to the common evaluation resolution before computing metrics. No method is given access to additional target-view images at test time.

\section{Additional View Synthesis Results}
\label{app:additional_view_results}

\subsection{In-Distribution Seen-Scenario Novel View Evaluation}
\label{app:indistribution_results}

\begin{table}[t]
\centering
\scriptsize
\caption{\textbf{In-distribution seen-scenario novel view evaluation.}
Following the seen-scenario split in Sec.~\ref{app:split_protocol}, we evaluate temporally held-out segments from scene UUIDs observed during training and report per-view generation fidelity across target cameras.}
\label{tab:wm_indistribution}
\setlength{\tabcolsep}{3.2pt}
\renewcommand{\arraystretch}{0.88}
\begin{adjustbox}{max width=0.98\columnwidth}
\begin{tabular}{@{}lccccc@{}}
\toprule
Target View
& PSNR $\uparrow$
& SSIM $\uparrow$
& LPIPS $\downarrow$
& FID $\downarrow$
& FVD $\downarrow$ \\
\midrule
Cross-left  & 15.33 & 0.563 & 0.505 & 127.86 & 127.58 \\
Cross-right & 14.80 & 0.526 & 0.521 & 133.95 & 123.24 \\
Rear-left   & 13.73 & 0.511 & 0.627 & 142.23 & 115.31 \\
Rear-right  & 12.68 & 0.462 & 0.625 & 149.98 & 121.56 \\
Rear-tele   & 13.18 & 0.450 & 0.637 & 159.80 & 117.31 \\
\bottomrule
\end{tabular}
\end{adjustbox}
\vspace{-2mm}
\end{table}

Table~\ref{tab:wm_indistribution} reports the in-distribution, seen-scenario view synthesis results. The cross-left and cross-right cameras obtain the strongest image-level fidelity, with higher PSNR and SSIM and lower LPIPS than the rear-facing cameras. This is expected because the cross views have stronger spatial overlap with the observed front camera, allowing the model to rely more directly on pose-grounded visual evidence. In contrast, the rear-left, rear-right, and rear-tele views require larger viewpoint extrapolation and contain regions that are weakly observed or unobserved from the front view, leading to lower image-level similarity scores.

Among the rear-facing views, rear-tele is particularly challenging because its narrower field of view and larger directional change make appearance reconstruction more sensitive to pose and temporal alignment errors. Nevertheless, the FVD scores remain relatively stable across target views, suggesting that the generated 41-frame clips preserve reasonable temporal coherence even when single-frame image fidelity decreases for more extrapolative cameras. Overall, the seen-scenario results show that OpenLongTail can synthesize temporally held-out target-rig views under familiar scene geometry, while also revealing the expected difficulty gap between partially overlapping cross views and more strongly extrapolative rear views.

\subsection{Additional Qualitative Results on Unseen Scenes}
\label{app:qual_unseen_scenes}

\begin{figure}[t]
\centering
\includegraphics[width=0.98\columnwidth]{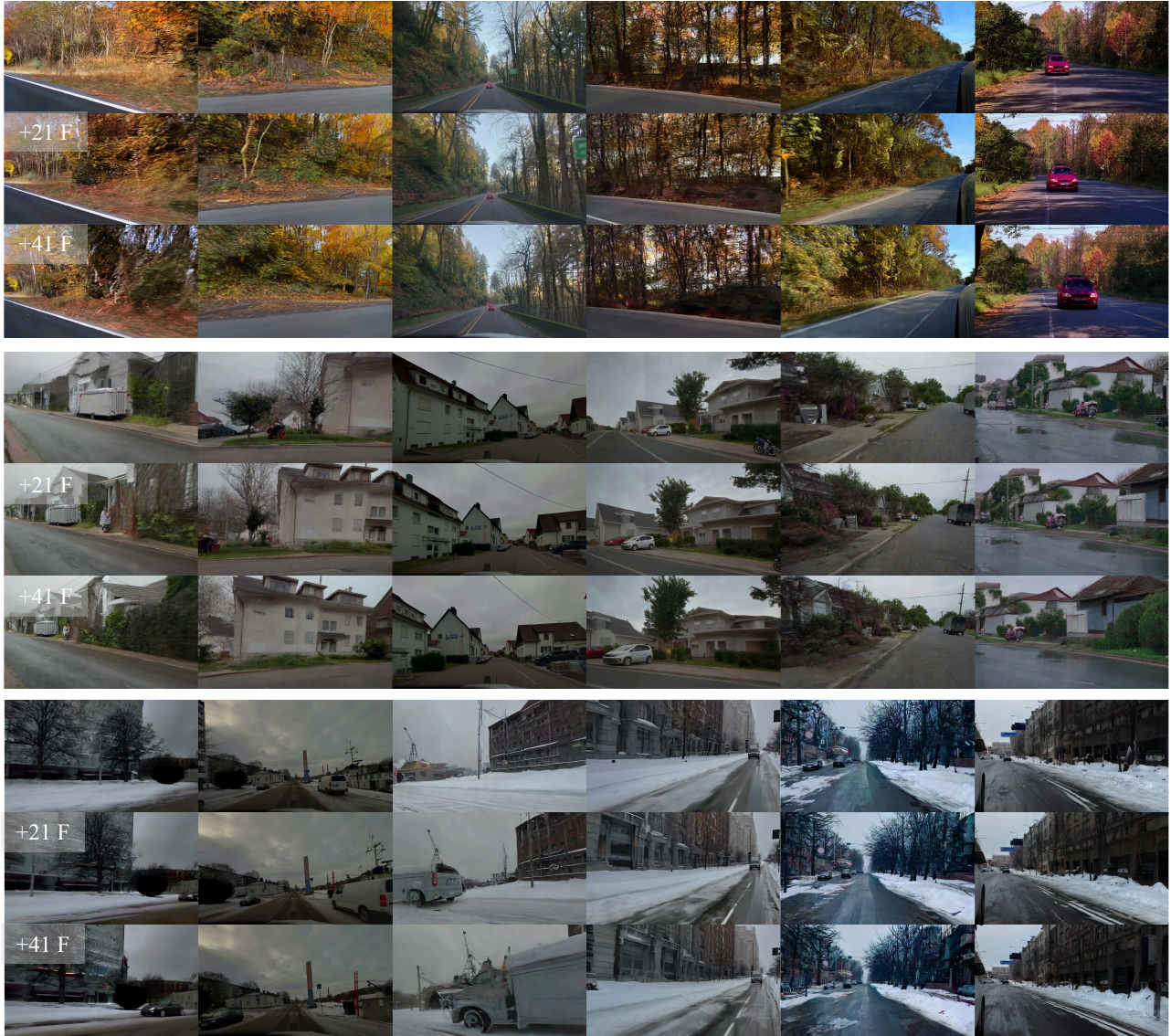}
\vspace{-2mm}
\caption{\textbf{Additional qualitative temporal novel view synthesis on unseen scenes.}
We show representative unseen-scene examples across diverse road layouts, appearance domains, and weather conditions, including rural tree-lined roads, residential neighborhoods, and snowy urban streets. Each example visualizes synthesized target-rig views over a 41-frame clip, with later rows corresponding to future frames such as the 21st frame and 41st frame.}
\label{fig:qual_unseen_scenes}
\vspace{-3mm}
\end{figure}

\begin{figure}[t]
\centering
\includegraphics[width=0.98\columnwidth]{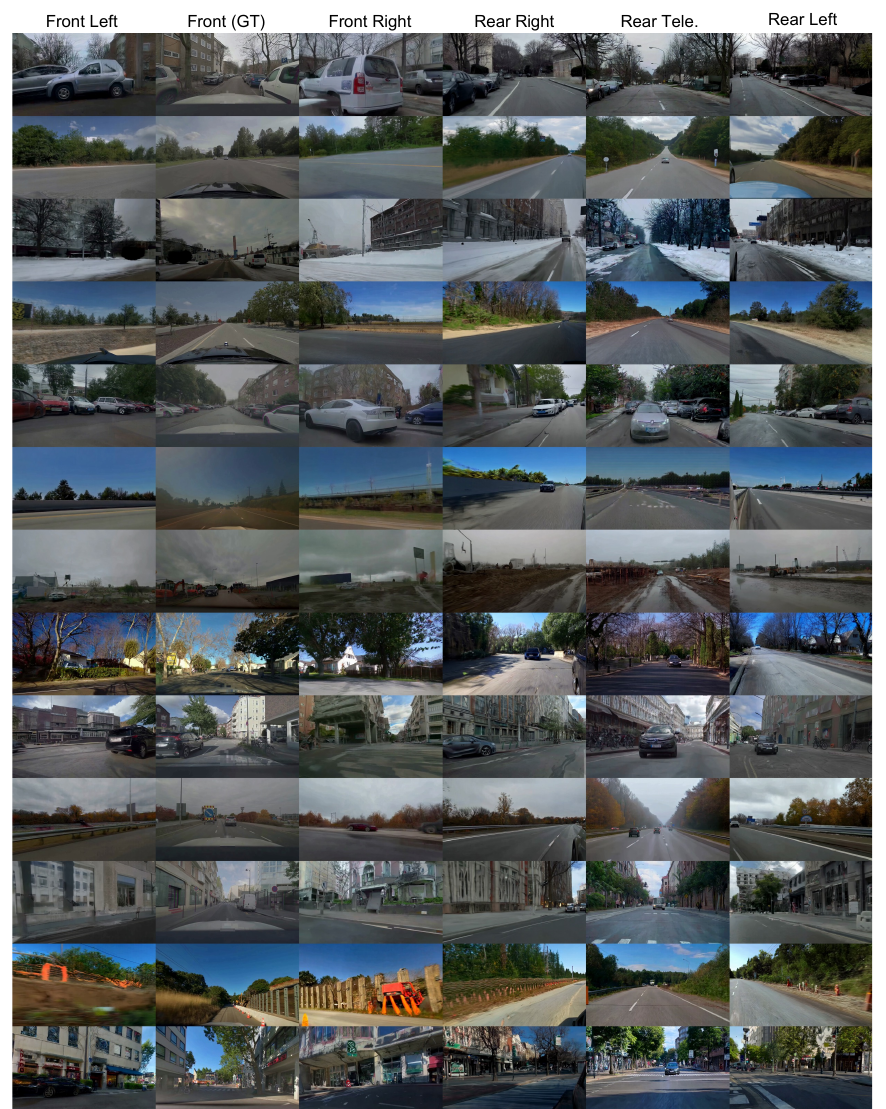}
\vspace{-2mm}
\caption{\textbf{Qualitative multi-view generation results on long-tail driving examples at the first frame.}
Each row corresponds to one data example, and each column shows one generated camera view in the order: left, front, right, rear-right, rear, and rear-left.}
\label{fig:qual_unseen_scenes_first}
\vspace{-3mm}
\end{figure}

\begin{figure}[t]
\centering
\includegraphics[width=0.98\columnwidth]{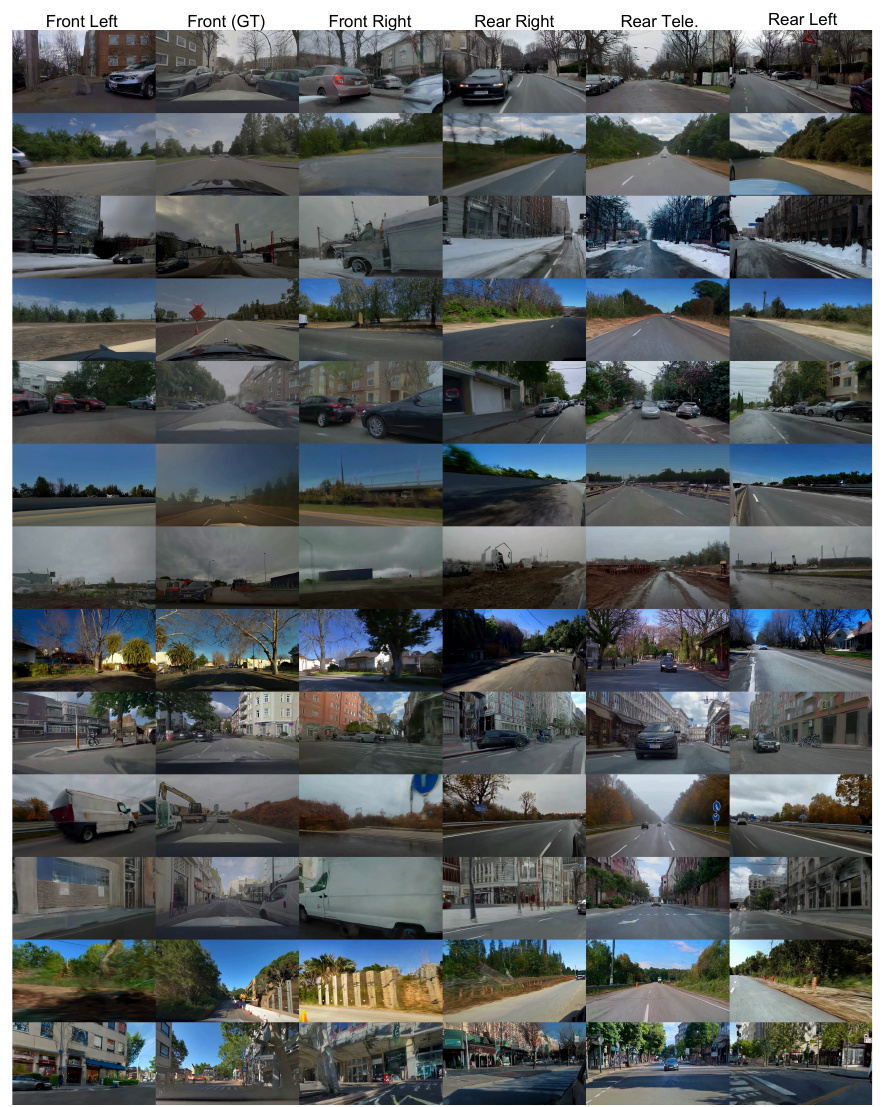}
\vspace{-2mm}
\caption{\textbf{Qualitative multi-view generation results on long-tail driving examples at the last frame.}
Each row corresponds to one data example, and each column shows one generated camera view in the order: left, front, right, rear-right, rear, and rear-left.}
\label{fig:qual_unseen_scenes_last}
\vspace{-3mm}
\end{figure}

Figure \ref{fig:qual_unseen_scenes}, Figure~\ref{fig:qual_unseen_scenes_first} and Figure \ref{fig:qual_unseen_scenes_last} provide additional qualitative results on unseen scenes. These examples are drawn from scene UUIDs that are not observed during training and therefore test whether the model can generalize beyond memorized scene geometry. The selected cases cover diverse visual domains, including an autumn rural road with dense vegetation, a residential neighborhood with buildings and parked vehicles, and a snowy urban street with low-contrast road boundaries. Across these settings, OpenLongTail produces plausible synchronized target-rig views and maintains the dominant scene structure, including road direction, roadside layout, buildings, vegetation, and surrounding traffic context.

The qualitative results also highlight the different difficulty levels across target views. Views with stronger geometric overlap to the observed front camera preserve sharper local appearance and more consistent road boundaries. More extrapolative rear-facing views require the model to hallucinate larger unobserved regions, making fine texture recovery harder. Nevertheless, the generated clips remain temporally coherent across the 41-frame horizon: static background elements move consistently with ego motion, road geometry remains stable, and large objects such as vehicles, houses, trees, and snowbanks do not exhibit severe frame-to-frame drift.

These unseen-scene examples complement the quantitative OOD results by showing that the model does not only improve image-level metrics, but also produces visually usable multi-view assets under meaningful domain shifts. In particular, the snowy and residential examples demonstrate robustness to appearance changes that are difficult to capture from front-view evidence alone. Overall, the qualitative results suggest that OpenLongTail can convert front-observed long-tail driving videos into coherent multi-camera clips suitable for downstream policy finetuning and evaluation.

\section{Pose Recovery Protocol}
\label{app:pose_protocol}

The pose recovery module used for data generation and evaluation follows the same protocol as the main paper. We use MapAnything to recover metric-scale ego trajectories from the input video and then apply a forward Kalman filter followed by a backward Rauch--Tung--Striebel smoother. The smoothed trajectory is used consistently for constructing Pl\"ucker-ray geometry, temporal depth warps, target-rig camera poses, and downstream policy inputs.

We report both metric-scale and Sim(3)-aligned pose evaluation. Metric-scale evaluation preserves the absolute scale of the recovered trajectory and directly measures whether the method provides metrically useful motion for view synthesis and policy learning. Sim(3)-aligned evaluation removes global rotation, translation, and scale before computing trajectory error, and therefore focuses on relative trajectory geometry. These two protocols measure different properties and should not be directly compared as the same metric.

\section{Downstream VLA Fine-Tuning Configuration}
\label{app:downstream_vla}

\subsection{Policy Backbone and Input Construction}
\label{app:vla_input}

We use Alpamayo-R1 \citep{wang2025alpamayo} as the downstream driving VLA backbone for supervised fine-tuning (SFT). Following the Alpamayo-R1 camera interface, each policy input is constructed from four camera views and the corresponding ego-motion signal. The front-view image is taken directly from the original video. The front-tele view is obtained by applying a deterministic FoV-Z crop to the front view with a $30^\circ$ field of view. The cross-left and cross-right views are either taken from the ground-truth multi-camera logs or synthesized by OpenLongTail under the target camera rig, depending on the SFT recipe. The ego-motion used by the policy is recovered by the same ego-trajectory module used for generation conditioning.

This protocol ensures that all SFT variants use the same policy architecture, camera interface, action format, and evaluation environment. The only difference across variants is the composition of additional training data and whether the non-front views are ground-truth multi-view observations or OpenLongTail-synthesized views.

\begin{table}[t]
\centering
\scriptsize
\caption{\textbf{Two-stage SFT configuration.}
Stage~1 trains the VLM with a trajectory-token prediction objective, while Stage~2 injects the Stage~1 VLM checkpoint and trains only the diffusion expert with the VLM frozen.}
\label{tab:two_stage_sft_config}
\setlength{\tabcolsep}{4.2pt}
\renewcommand{\arraystretch}{0.95}
\begin{adjustbox}{max width=\linewidth}
\begin{tabular}{@{}lcc@{}}
\toprule
\textbf{Parameter} & \textbf{Stage 1} & \textbf{Stage 2} \\
\midrule
Backbone
& Alpamayo-R1-10B
& Alpamayo-R1-10B + Stage~1 VLM replacement \\
Config yaml
& \texttt{sft\_stage1}
& \texttt{sft\_stage2} \\
Loss
& Trajectory-token CE
& Flow-matching \\
Trainable parameters
& Full VLM
& Diffusion expert only \\
VLM cotraining
& --
& \texttt{cotrain\_vlm=false} \\
Stage~1 checkpoint injection
& --
& \texttt{model.stage1\_vlm\_checkpoint\_path} \\
Epochs
& 3
& 3 \\
Total steps
& 1,149
& 4,593 \\
Steps per epoch
& 383
& 1,531 \\
GPUs
& 8$\times$ H200
& 8$\times$ H200 \\
Per-device train batch size
& 1
& 1 \\
Gradient accumulation steps
& 4
& 1 \\
Global batch size
& $8 \times 1 \times 4 = 32$
& $8 \times 1 \times 1 = 8$ \\
Samples per epoch
& $\approx$12,256
& $\approx$12,248 \\
Precision
& bf16
& bf16 \\
DeepSpeed
& ZeRO-2
& none \\
Gradient checkpointing
& \checkmark
& -- \\
Learning rate
& $1\times10^{-5}$
& $1\times10^{-4}$ \\
LR scheduler
& cosine warmup with min LR
& cosine warmup with min LR \\
Minimum LR
& $1\times10^{-6}$
& $1\times10^{-6}$ \\
Warmup steps
& 500
& 100 \\
\bottomrule
\end{tabular}
\end{adjustbox}
\vspace{-1mm}
\end{table}

We use a two-stage SFT strategy for Alpamayo-R1-10B \citep{wang2025alpamayo}. Stage~1 trains the full VLM with a trajectory-token CE loss using \texttt{sft\_stage1} for 3 epochs, totaling 1,149 steps with a global batch size of 32 on 8 H200 GPUs. Stage~2 loads the Stage~1 VLM checkpoint via \texttt{model.stage1\_vlm\_checkpoint\_path}, freezes the VLM with \texttt{cotrain\_vlm=false}, and trains only the diffusion expert with a flow-matching loss using \texttt{sft\_stage2}. This stage runs for 3 epochs and 4,593 steps with a global batch size of 8. Both stages use bf16 precision and a cosine warmup schedule with minimum learning rate $1\times10^{-6}$; Stage~1 uses $1\times10^{-5}$ learning rate with ZeRO-2 and gradient checkpointing, while Stage~2 uses $1\times10^{-4}$ without DeepSpeed.

\subsection{SFT Data Composition}
\label{app:vla_data_composition}

We fine-tune Alpamayo-R1 for three epochs for each SFT recipe. All recipes use the same optimization setup and differ only in the additional training data mixture. The nominal data consists of 10K randomly sampled driving trajectories. The long-tail data includes PAV \citep{nvidia_physicalai_av_2025} long-tail clips and Waymo E2E \citep{xu2025wod} clips. For each long-tail source, we consider both ground-truth multi-view data, when available, and OpenLongTail-synthesized multi-view assets.

\begin{table}[t]
\centering
\scriptsize
\caption{\textbf{Downstream VLA SFT data composition.}
All variants fine-tune Alpamayo-R1 for three epochs under the same optimization and camera-input protocol. The recipes differ only in the additional training data used for SFT. GT denotes ground-truth multi-view data, while Syn denotes OpenLongTail-synthesized multi-view assets.}
\label{tab:vla_sft_data_composition}
\setlength{\tabcolsep}{3.2pt}
\renewcommand{\arraystretch}{0.92}
\begin{adjustbox}{max width=\textwidth}
\begin{tabular}{@{}lccccc l@{}}
\toprule
Recipe
& 10K Rand
& NV-OOD GT
& NV-OOD Syn
& Waymo-E2E GT
& Waymo-E2E Syn
& Purpose \\
\midrule
Base Alpamayo-R1
&  &  &  &  &
& no additional SFT baseline \\
Base Alpamayo 1.5
&  &  &  &  &
& no additional SFT baseline \\
\midrule

10K nominal SFT
& \checkmark &  &  &  &
& nominal-data adaptation baseline \\
NV-OOD GT SFT
& \checkmark & \checkmark &  &  &
& ground-truth NV long-tail upper-bound comparison \\
NV-OOD Syn SFT
& \checkmark &  & \checkmark &  &
& evaluates synthesized NV long-tail assets from PAV/NV source \\
GT long-tail mixture
& \checkmark & \checkmark &  & \checkmark &
& evaluates NV+Waymo ground-truth long-tail SFT \\
Syn long-tail mixture
& \checkmark &  & \checkmark &  & \checkmark
& evaluates NV+Waymo OpenLongTail-generated SFT \\
\bottomrule
\end{tabular}
\end{adjustbox}
\vspace{-2mm}
\end{table}

For GT recipes, the target views are obtained from synchronized calibrated multi-camera logs. For Syn recipes, the target views are synthesized by OpenLongTail from the corresponding front-view videos using the same target-rig camera protocol. This design isolates the effect of synthesized multi-view long-tail assets while keeping the policy interface and SFT procedure fixed.

\subsection{AlpaSim Closed-Loop Evaluation}
\label{app:alphasim_eval}

\paragraph{AlpaSim Simulation Environment.}
We use AlpaSim as a photorealistic closed-loop simulation environment for evaluating end-to-end autonomous driving policies. Unlike open-loop evaluation, where the policy is tested against fixed logged observations, AlpaSim re-renders sensor inputs from the policy’s current simulated ego state at every rollout step. As a result, the policy observes the consequences of its own actions, enabling evaluation of compounding errors, recovery behavior, off-road motion, wrong-lane driving, collisions, and route progress under realistic visual conditions.

\paragraph{Scenario Asset Construction.}
AlpaSim converts real-world driving logs into reusable simulation assets. Starting from synchronized multi-camera data, calibration, ego poses, object tracks, and map information, the pipeline first normalizes all sensors and trajectories into a shared world coordinate system. NuRec is then used to reconstruct the scene as a neural 3D asset, typically represented with 3D Gaussian Splatting. This representation preserves high-fidelity appearance such as road texture, lane markings, vegetation, buildings, signs, and construction elements, while supporting efficient novel-view rendering from ego poses that differ from the original logged trajectory.

Beyond the visual 3DGS scene, each asset also contains explicit driving structure, including a ground mesh, drivable surface, lane geometry, road boundaries, route information, and dynamic-agent tracks. The ground mesh and map layers provide the simulator with interpretable geometry for collision checking, off-road detection, lane compliance, and progress measurement, while the neural reconstruction provides realistic camera observations for the policy. This separation allows AlpaSim to combine photorealistic rendering with reliable closed-loop evaluation metrics.

\paragraph{Closed-Loop Rendering and Novel-View Synthesis.}
During simulation, AlpaSim maintains the ego vehicle state, places the virtual camera rig at the corresponding pose, and queries the neural renderer to synthesize the current multi-camera observations. The driving policy receives these rendered images and outputs an action or future trajectory. The simulator then updates the ego state through a vehicle dynamics or trajectory-following model and repeats the process. Because the next observation depends on the policy’s previous action, AlpaSim is not simple dataset replay; it is a closed-loop environment where visual feedback changes as the policy deviates from trajectory.

Novel-view synthesis is the key mechanism that enables this feedback loop. Since the ego vehicle may move to poses not observed in the original log, AlpaSim uses the reconstructed 3DGS scene to render photorealistic camera views from arbitrary nearby viewpoints. This is particularly important for evaluating VLA driving models, which rely on image-level cues such as lane markings, cones, signs, workers, police officers, occlusions, and irregular intersection layouts. Therefore, AlpaSim provides a realistic downstream testbed for measuring whether generated long-tail multi-view data improves actual closed-loop driving behavior.



\clearpage

%% file: example.bib
@inproceedings{ettinger2021large,
  title={Large scale interactive motion forecasting for autonomous driving: The waymo open motion dataset},
  author={Ettinger, Scott and Cheng, Shuyang and Caine, Benjamin and Liu, Chenxi and Zhao, Hang and Pradhan, Sabeek and Chai, Yuning and Sapp, Ben and Qi, Charles R and Zhou, Yin and others},
  booktitle={Proceedings of the IEEE/CVF international conference on computer vision},
  pages={9710--9719},
  year={2021}
}

@misc{nexar2025dashcamcollisionprediction,
      title={Nexar Dashcam Collision Prediction Dataset and Challenge}, 
      author={Daniel C. Moura and Shizhan Zhu and Orly Zvitia},
      year={2025},
      eprint={2503.03848},
      archivePrefix={arXiv},
      primaryClass={cs.CV},
      url={https://arxiv.org/abs/2503.03848}, 
}

@article{gao2026steervla,
  title={SteerVLA: Steering Vision-Language-Action Models in Long-Tail Driving Scenarios},
  author={Gao, Tian and Tan, Celine and Glossop, Catherine and Gao, Timothy and Sun, Jiankai and Stachowicz, Kyle and Wu, Shirley and Mees, Oier and Sadigh, Dorsa and Levine, Sergey and others},
  journal={arXiv preprint arXiv:2602.08440},
  year={2026}
}

@inproceedings{wu2025cat4d,
  title={Cat4d: Create anything in 4d with multi-view video diffusion models},
  author={Wu, Rundi and Gao, Ruiqi and Poole, Ben and Trevithick, Alex and Zheng, Changxi and Barron, Jonathan T and Holynski, Aleksander},
  booktitle={Proceedings of the IEEE/CVF Conference on Computer Vision and Pattern Recognition},
  pages={26057--26068},
  year={2025}
}

@article{van2026anyview,
  title={AnyView: Synthesizing Any Novel View in Dynamic Scenes},
  author={Van Hoorick, Basile and Chen, Dian and Iwase, Shun and Tokmakov, Pavel and Irshad, Muhammad Zubair and Vasiljevic, Igor and Gupta, Swati and Cheng, Fangzhou and Zakharov, Sergey and Guizilini, Vitor Campagnolo},
  journal={arXiv preprint arXiv:2601.16982},
  year={2026}
}

@article{yu2024viewcrafter,
  title={Viewcrafter: Taming video diffusion models for high-fidelity novel view synthesis},
  author={Yu, Wangbo and Xing, Jinbo and Yuan, Li and Hu, Wenbo and Li, Xiaoyu and Huang, Zhipeng and Gao, Xiangjun and Wong, Tien-Tsin and Shan, Ying and Tian, Yonghong},
  journal={arXiv preprint arXiv:2409.02048},
  year={2024}
}

@inproceedings{zhao2025drivedreamer,
  title={Drivedreamer-2: Llm-enhanced world models for diverse driving video generation},
  author={Zhao, Guosheng and Wang, Xiaofeng and Zhu, Zheng and Chen, Xinze and Huang, Guan and Bao, Xiaoyi and Wang, Xingang},
  booktitle={Proceedings of the AAAI Conference on Artificial Intelligence},
  volume={39},
  number={10},
  pages={10412--10420},
  year={2025}
}

@article{russell2025gaia,
  title={Gaia-2: A controllable multi-view generative world model for autonomous driving},
  author={Russell, Lloyd and Hu, Anthony and Bertoni, Lorenzo and Fedoseev, George and Shotton, Jamie and Arani, Elahe and Corrado, Gianluca},
  journal={arXiv preprint arXiv:2503.20523},
  year={2025}
}

@inproceedings{wang2024driving,
  title={Driving into the future: Multiview visual forecasting and planning with world model for autonomous driving},
  author={Wang, Yuqi and He, Jiawei and Fan, Lue and Li, Hongxin and Chen, Yuntao and Zhang, Zhaoxiang},
  booktitle={Proceedings of the IEEE/CVF Conference on Computer Vision and Pattern Recognition},
  pages={14749--14759},
  year={2024}
}

@article{li2026droid,
  title={DROID-SLAM in the Wild},
  author={Li, Moyang and Zhu, Zihan and Pollefeys, Marc and Barath, Daniel},
  journal={arXiv preprint arXiv:2603.19076},
  year={2026}
}

@misc{nvidia_physicalai_av_2025,
  author       = {{NVIDIA Corporation}},
  title        = {{PhysicalAI-Autonomous-Vehicles}},
  year         = {2025},
  howpublished = {\url{https://huggingface.co/datasets/nvidia/PhysicalAI-Autonomous-Vehicles}},
}

@inproceedings{xiao2021pandaset,
  title={Pandaset: Advanced sensor suite dataset for autonomous driving},
  author={Xiao, Pengchuan and Shao, Zhenlei and Hao, Steven and Zhang, Zishuo and Chai, Xiaolin and Jiao, Judy and Li, Zesong and Wu, Jian and Sun, Kai and Jiang, Kun and others},
  booktitle={2021 IEEE international intelligent transportation systems conference (ITSC)},
  pages={3095--3101},
  year={2021},
  organization={IEEE}
}

@software{alpasim_2025,
  author       = {
    NVIDIA and
    Yulong Cao and
    Riccardo de Lutio and
    Sanja Fidler and
    Guillermo Garcia Cobo and
    Zan Gojcic and
    Maximilian Igl and
    Boris Ivanovic and
    Peter Karkus and
    Janick Martinez Esturo and
    Marco Pavone and
    Aaron Smith and
    Ellie Tanimura and
    Michal Tyszkiewicz and
    Michael Watson and
    Qi Wu and
    Le Zhang
  },
  title        = {AlpaSim: A Modular, Lightweight, and Data-Driven Research Simulator for Autonomous Driving},
  year         = {2025},
  month        = {October},
  url          = {https://github.com/NVlabs/alpasim},
}

@article{unterthiner2018towards,
  title={Towards accurate generative models of video: A new metric \& challenges},
  author={Unterthiner, Thomas and Van Steenkiste, Sjoerd and Kurach, Karol and Marinier, Raphael and Michalski, Marcin and Gelly, Sylvain},
  journal={arXiv preprint arXiv:1812.01717},
  year={2018}
}

@article{wang2004image,
  title={Image quality assessment: from error visibility to structural similarity},
  author={Wang, Zhou and Bovik, Alan C and Sheikh, Hamid R and Simoncelli, Eero P},
  journal={IEEE transactions on image processing},
  volume={13},
  number={4},
  pages={600--612},
  year={2004},
  publisher={IEEE}
}

@article{heusel2017gans,
  title={Gans trained by a two time-scale update rule converge to a local nash equilibrium},
  author={Heusel, Martin and Ramsauer, Hubert and Unterthiner, Thomas and Nessler, Bernhard and Hochreiter, Sepp},
  journal={Advances in neural information processing systems},
  volume={30},
  year={2017}
}

@inproceedings{zhang2018unreasonable,
  title={The unreasonable effectiveness of deep features as a perceptual metric},
  author={Zhang, Richard and Isola, Phillip and Efros, Alexei A and Shechtman, Eli and Wang, Oliver},
  booktitle={Proceedings of the IEEE conference on computer vision and pattern recognition},
  pages={586--595},
  year={2018}
}

@inproceedings{li2025megasam,
  title={Megasam: Accurate, fast and robust structure and motion from casual dynamic videos},
  author={Li, Zhengqi and Tucker, Richard and Cole, Forrester and Wang, Qianqian and Jin, Linyi and Ye, Vickie and Kanazawa, Angjoo and Holynski, Aleksander and Snavely, Noah},
  booktitle={Proceedings of the IEEE/CVF Conference on Computer Vision and Pattern Recognition},
  pages={10486--10496},
  year={2025}
}

@inproceedings{lin2026vista4d,
    author = {Lin, {Kuan Heng} and Liu, Zhizheng and Salamanca, Pablo and Kant, Yash and Burgert, Ryan and Xu, Yuancheng and Namekata, Koichi and Zhao, Yiwei and Zhou, Bolei and Goldblum, Micah and Debevec, Paul and Yu, Ning},
    booktitle = {Proceedings of the IEEE/CVF Conference on Computer Vision and Pattern Recognition},
    title = {{Vista4D}: Video Reshooting with 4D Point Clouds},
    year = {2026}
}

@inproceedings{ren2025gen3c,
  title={GEN3C: 3D-Informed World-Consistent Video Generation with Precise Camera Control},
  author={Ren, Xuanchi and Shen, Tianchang and Huang, Jiahui and Ling, Huan and Lu, Yifan and Nimier-David, Merlin and M\"uller, Thomas and Keller, Alexander and Fidler, Sanja and Gao, Jun},
  booktitle={Proceedings of the IEEE/CVF Conference on Computer Vision and Pattern Recognition},
  year={2025}
}

@article{mao2021one,
  title={One million scenes for autonomous driving: Once dataset},
  author={Mao, Jiageng and Niu, Minzhe and Jiang, Chenhan and Liang, Hanxue and Chen, Jingheng and Liang, Xiaodan and Li, Yamin and Ye, Chaoqiang and Zhang, Wei and Li, Zhenguo and others},
  journal={arXiv preprint arXiv:2106.11037},
  year={2021}
}

@inproceedings{yu2020bdd100k,
  title={Bdd100k: A diverse driving dataset for heterogeneous multitask learning},
  author={Yu, Fisher and Chen, Haofeng and Wang, Xin and Xian, Wenqi and Chen, Yingying and Liu, Fangchen and Madhavan, Vashisht and Darrell, Trevor},
  booktitle={Proceedings of the IEEE/CVF conference on computer vision and pattern recognition},
  pages={2636--2645},
  year={2020}
}

@article{wilson2023argoverse,
  title={Argoverse 2: Next generation datasets for self-driving perception and forecasting},
  author={Wilson, Benjamin and Qi, William and Agarwal, Tanmay and Lambert, John and Singh, Jagjeet and Khandelwal, Siddhesh and Pan, Bowen and Kumar, Ratnesh and Hartnett, Andrew and Pontes, Jhony Kaesemodel and others},
  journal={arXiv preprint arXiv:2301.00493},
  year={2023}
}

@inproceedings{huang2025vipe,
    title={ViPE: Video Pose Engine for 3D Geometric Perception},
    author={Huang, Jiahui and Zhou, Qunjie and Rabeti, Hesam and Korovko, Aleksandr and Ling, Huan and Ren, Xuanchi and Shen, Tianchang and Gao, Jun and Slepichev, Dmitry and Lin, Chen-Hsuan and Ren, Jiawei and Xie, Kevin and Biswas, Joydeep and Leal-Taixe, Laura and Fidler, Sanja},
    booktitle={NVIDIA Research Whitepapers arXiv:2508.10934},
    year={2025}
}

@inproceedings{sun2020scalability,
  title={Scalability in perception for autonomous driving: Waymo open dataset},
  author={Sun, Pei and Kretzschmar, Henrik and Dotiwalla, Xerxes and Chouard, Aurelien and Patnaik, Vijaysai and Tsui, Paul and Guo, James and Zhou, Yin and Chai, Yuning and Caine, Benjamin and others},
  booktitle={Proceedings of the IEEE/CVF conference on computer vision and pattern recognition},
  pages={2446--2454},
  year={2020}
}

@article{li2022coda,
title={CODA: A Real-World Road Corner Case Dataset for Object Detection in Autonomous Driving},
author={Li, Kaican and Chen, Kai and Wang, Haoyu and Hong, Lanqing and Ye, Chaoqiang and Han, Jianhua and Chen, Yukuai and Zhang, Wei and Xu, Chunjing and Yeung, Dit-Yan and others},
journal={arXiv preprint arXiv:2203.07724},
year={2022}
}

@inproceedings{caesar2020nuscenes,
  title={nuscenes: A multimodal dataset for autonomous driving},
  author={Caesar, Holger and Bankiti, Varun and Lang, Alex H and Vora, Sourabh and Liong, Venice Erin and Xu, Qiang and Krishnan, Anush and Pan, Yu and Baldan, Giancarlo and Beijbom, Oscar},
  booktitle={Proceedings of the IEEE/CVF conference on computer vision and pattern recognition},
  pages={11621--11631},
  year={2020}
}

@article{xu2025wod,
  title={Wod-e2e: Waymo open dataset for end-to-end driving in challenging long-tail scenarios},
  author={Xu, Runsheng and Lin, Hubert and Jeon, Wonseok and Feng, Hao and Zou, Yuliang and Sun, Liting and Gorman, John and Tolstaya, Ekaterina and Tang, Sarah and White, Brandyn and others},
  journal={arXiv preprint arXiv:2510.26125},
  year={2025}
}

@article{wan2025wan,
  title={Wan: Open and advanced large-scale video generative models},
  author={Wan, Team and Wang, Ang and Ai, Baole and Wen, Bin and Mao, Chaojie and Xie, Chen-Wei and Chen, Di and Yu, Feiwu and Zhao, Haiming and Yang, Jianxiao and others},
  journal={arXiv preprint arXiv:2503.20314},
  year={2025}
}

@article{yang2024cogvideox,
  title={Cogvideox: Text-to-video diffusion models with an expert transformer},
  author={Yang, Zhuoyi and Teng, Jiayan and Zheng, Wendi and Ding, Ming and Huang, Shiyu and Xu, Jiazheng and Yang, Yuanming and Hong, Wenyi and Zhang, Xiaohan and Feng, Guanyu and others},
  journal={arXiv preprint arXiv:2408.06072},
  year={2024}
}

@article{hu2023gaia,
  title={Gaia-1: A generative world model for autonomous driving},
  author={Hu, Anthony and Russell, Lloyd and Yeo, Hudson and Murez, Zak and Fedoseev, George and Kendall, Alex and Shotton, Jamie and Corrado, Gianluca},
  journal={arXiv preprint arXiv:2309.17080},
  year={2023}
}

@article{chen2025skyreels,
  title={Skyreels-v2: Infinite-length film generative model},
  author={Chen, Guibin and Lin, Dixuan and Yang, Jiangping and Lin, Chunze and Zhu, Junchen and Fan, Mingyuan and Zhang, Hao and Chen, Sheng and Chen, Zheng and Ma, Chengcheng and others},
  journal={arXiv preprint arXiv:2504.13074},
  year={2025}
}

@article{ali2025world,
  title={World simulation with video foundation models for physical ai},
  author={Ali, Arslan and Bai, Junjie and Bala, Maciej and Balaji, Yogesh and Blakeman, Aaron and Cai, Tiffany and Cao, Jiaxin and Cao, Tianshi and Cha, Elizabeth and Chao, Yu-Wei and others},
  journal={arXiv preprint arXiv:2511.00062},
  year={2025}
}

@inproceedings{wang2025vggt,
  title={Vggt: Visual geometry grounded transformer},
  author={Wang, Jianyuan and Chen, Minghao and Karaev, Nikita and Vedaldi, Andrea and Rupprecht, Christian and Novotny, David},
  booktitle={Proceedings of the Computer Vision and Pattern Recognition Conference},
  pages={5294--5306},
  year={2025}
}

@inproceedings{wang2024drivedreamer,
  title={Drivedreamer: Towards real-world-drive world models for autonomous driving},
  author={Wang, Xiaofeng and Zhu, Zheng and Huang, Guan and Chen, Xinze and Zhu, Jiagang and Lu, Jiwen},
  booktitle={European conference on computer vision},
  pages={55--72},
  year={2024},
  organization={Springer}
}

@article{gao2023magicdrive,
  title={Magicdrive: Street view generation with diverse 3d geometry control},
  author={Gao, Ruiyuan and Chen, Kai and Xie, Enze and Hong, Lanqing and Li, Zhenguo and Yeung, Dit-Yan and Xu, Qiang},
  journal={arXiv preprint arXiv:2310.02601},
  year={2023}
}

@article{keetha2025mapanything,
  title={Mapanything: Universal feed-forward metric 3d reconstruction},
  author={Keetha, Nikhil and M{\"u}ller, Norman and Sch{\"o}nberger, Johannes and Porzi, Lorenzo and Zhang, Yuchen and Fischer, Tobias and Knapitsch, Arno and Zauss, Duncan and Weber, Ethan and Antunes, Nelson and others},
  journal={arXiv preprint arXiv:2509.13414},
  year={2025}
}

@article{yang2026neoverse,
  title={NeoVerse: Enhancing 4D World Model with in-the-wild Monocular Videos},
  author={Yang, Yuxue and Fan, Lue and Shi, Ziqi and Peng, Junran and Wang, Feng and Zhang, Zhaoxiang},
  journal={arXiv preprint arXiv:2601.00393},
  year={2026}
}

@article{ren2025cosmos,
  title={Cosmos-drive-dreams: Scalable synthetic driving data generation with world foundation models},
  author={Ren, Xuanchi and Lu, Yifan and Cao, Tianshi and Gao, Ruiyuan and Huang, Shengyu and Sabour, Amirmojtaba and Shen, Tianchang and Pfaff, Tobias and Wu, Jay Zhangjie and Chen, Runjian and others},
  journal={arXiv preprint arXiv:2506.09042},
  year={2025}
}

@article{wang2025alpamayo,
  title={Alpamayo-r1: Bridging reasoning and action prediction for generalizable autonomous driving in the long tail},
  author={Wang, Yan and Luo, Wenjie and Bai, Junjie and Cao, Yulong and Che, Tong and Chen, Ke and Chen, Yuxiao and Diamond, Jenna and Ding, Yifan and Ding, Wenhao and others},
  journal={arXiv preprint arXiv:2511.00088},
  year={2025}
}

@inproceedings{bai2025recammaster,
  title={Recammaster: Camera-controlled generative rendering from a single video},
  author={Bai, Jianhong and Xia, Menghan and Fu, Xiao and Wang, Xintao and Mu, Lianrui and Cao, Jinwen and Liu, Zuozhu and Hu, Haoji and Bai, Xiang and Wan, Pengfei and others},
  booktitle={Proceedings of the IEEE/CVF International Conference on Computer Vision},
  pages={14834--14844},
  year={2025}
}

@inproceedings{yu2025trajectorycrafter,
  title={Trajectorycrafter: Redirecting camera trajectory for monocular videos via diffusion models},
  author={Yu, Mark and Hu, Wenbo and Xing, Jinbo and Shan, Ying},
  booktitle={Proceedings of the IEEE/CVF international conference on computer vision},
  pages={100--111},
  year={2025}
}

@article{li2025drivevla,
  title={DriveVLA-W0: World models amplify data scaling law in autonomous driving},
  author={Li, Yingyan and Shang, Shuyao and Liu, Weisong and Zhan, Bing and Wang, Haochen and Wang, Yuqi and Chen, Yuntao and Wang, Xiaoman and An, Yasong and Tang, Chufeng and others},
  journal={arXiv preprint arXiv:2510.12796},
  year={2025}
}

@article{jiang2025alphadrive,
  title={Alphadrive: Unleashing the power of vlms in autonomous driving via reinforcement learning and reasoning},
  author={Jiang, Bo and Chen, Shaoyu and Zhang, Qian and Liu, Wenyu and Wang, Xinggang},
  journal={arXiv preprint arXiv:2503.07608},
  year={2025}
}

@article{xu2024vlm,
  title={Vlm-ad: End-to-end autonomous driving through vision-language model supervision},
  author={Xu, Yi and Hu, Yuxin and Zhang, Zaiwei and Meyer, Gregory P and Mustikovela, Siva Karthik and Srinivasa, Siddhartha and Wolff, Eric M and Huang, Xin},
  journal={arXiv preprint arXiv:2412.14446},
  year={2024}
}

@inproceedings{zhou2026opendrivevla,
  title={Opendrivevla: Towards end-to-end autonomous driving with large vision language action model},
  author={Zhou, Xingcheng and Han, Xuyuan and Yang, Feng and Ma, Yunpu and Tresp, Volker and Knoll, Alois},
  booktitle={Proceedings of the AAAI Conference on Artificial Intelligence},
  volume={40},
  number={16},
  pages={13782--13790},
  year={2026}
}

@article{zhou2025autovla,
  title={Autovla: A vision-language-action model for end-to-end autonomous driving with adaptive reasoning and reinforcement fine-tuning},
  author={Zhou, Zewei and Cai, Tianhui and Zhao, Seth Z and Zhang, Yun and Huang, Zhiyu and Zhou, Bolei and Ma, Jiaqi},
  journal={arXiv preprint arXiv:2506.13757},
  year={2025}
}

@article{yuan2025autodrive,
  title={AutoDrive-R$^2$: Incentivizing Reasoning and Self-Reflection Capacity for VLA Model in Autonomous Driving},
  author={Yuan, Zhenlong and Qian, Chengxuan and Tang, Jing and Chen, Rui and Song, Zijian and Sun, Lei and Chu, Xiangxiang and Cai, Yujun and Zhang, Dapeng and Li, Shuo},
  journal={arXiv preprint arXiv:2509.01944},
  year={2025}
}

@inproceedings{sima2024drivelm,
  title={Drivelm: Driving with graph visual question answering},
  author={Sima, Chonghao and Renz, Katrin and Chitta, Kashyap and Chen, Li and Zhang, Hanxue and Xie, Chengen and Bei{\ss}wenger, Jens and Luo, Ping and Geiger, Andreas and Li, Hongyang},
  booktitle={European conference on computer vision},
  pages={256--274},
  year={2024},
  organization={Springer}
}

@inproceedings{shao2024lmdrive,
  title={Lmdrive: Closed-loop end-to-end driving with large language models},
  author={Shao, Hao and Hu, Yuxuan and Wang, Letian and Song, Guanglu and Waslander, Steven L and Liu, Yu and Li, Hongsheng},
  booktitle={Proceedings of the IEEE/CVF conference on computer vision and pattern recognition},
  pages={15120--15130},
  year={2024}
}

@article{xu2024drivegpt4,
  title={Drivegpt4: Interpretable end-to-end autonomous driving via large language model},
  author={Xu, Zhenhua and Zhang, Yujia and Xie, Enze and Zhao, Zhen and Guo, Yong and Wong, Kwan-Yee K and Li, Zhenguo and Zhao, Hengshuang},
  journal={IEEE Robotics and Automation Letters},
  volume={9},
  number={10},
  pages={8186--8193},
  year={2024},
  publisher={IEEE}
}

@article{wang2023drivemlm,
  title={Drivemlm: Aligning multi-modal large language models with behavioral planning states for autonomous driving},
  author={Wang, Wenhai and Xie, Jiangwei and Hu, ChuanYang and Zou, Haoming and Fan, Jianan and Tong, Wenwen and Wen, Yang and Wu, Silei and Deng, Hanming and Li, Zhiqi and others},
  journal={arXiv preprint arXiv:2312.09245},
  year={2023}
}

@inproceedings{ma2024dolphins,
  title={Dolphins: Multimodal language model for driving},
  author={Ma, Yingzi and Cao, Yulong and Sun, Jiachen and Pavone, Marco and Xiao, Chaowei},
  booktitle={European Conference on Computer Vision},
  pages={403--420},
  year={2024},
  organization={Springer}
}

@article{hwang2024emma,
  title={Emma: End-to-end multimodal model for autonomous driving},
  author={Hwang, Jyh-Jing and Xu, Runsheng and Lin, Hubert and Hung, Wei-Chih and Ji, Jingwei and Choi, Kristy and Huang, Di and He, Tong and Covington, Paul and Sapp, Benjamin and others},
  journal={arXiv preprint arXiv:2410.23262},
  year={2024}
}

@article{jiang2024senna,
  title={Senna: Bridging large vision-language models and end-to-end autonomous driving},
  author={Jiang, Bo and Chen, Shaoyu and Liao, Bencheng and Zhang, Xingyu and Yin, Wei and Zhang, Qian and Huang, Chang and Liu, Wenyu and Wang, Xinggang},
  journal={arXiv preprint arXiv:2410.22313},
  year={2024}
}

@article{feng2025verdi,
  title={Verdi: Vlm-embedded reasoning for autonomous driving},
  author={Feng, Bowen and Mei, Zhiting and Ost, Julian and Ghilotti, Filippo and Li, Baiang and Girgis, Roger and Majumdar, Anirudha and Heide, Felix},
  journal={arXiv preprint arXiv:2505.15925},
  year={2025}
}

@article{jiang2025diffvla,
  title={Diffvla: Vision-language guided diffusion planning for autonomous driving},
  author={Jiang, Anqing and Gao, Yu and Sun, Zhigang and Wang, Yiru and Wang, Jijun and Chai, Jinghao and Cao, Qian and Heng, Yuweng and Jiang, Hao and Dong, Yunda and others},
  journal={arXiv preprint arXiv:2505.19381},
  year={2025}
}

@article{rowe2025poutine,
  title={Poutine: Vision-language-trajectory pre-training and reinforcement learning post-training enable robust end-to-end autonomous driving},
  author={Rowe, Luke and de Schaetzen, Rodrigue and Girgis, Roger and Pal, Christopher and Paull, Liam},
  journal={arXiv preprint arXiv:2506.11234},
  year={2025}
}

@article{song2025lmad,
  title={LMAD: Integrated End-to-End Vision-Language Model for Explainable Autonomous Driving},
  author={Song, Nan and Zhang, Bozhou and Zhu, Xiatian and Deng, Jiankang and Zhang, Li},
  journal={arXiv preprint arXiv:2508.12404},
  year={2025}
}

@inproceedings{xie2025vlms,
  title={Are VLMs Ready for Autonomous Driving? An Empirical Study from the Reliability, Data and Metric Perspectives},
  author={Xie, Shaoyuan and Kong, Lingdong and Dong, Yuhao and Sima, Chonghao and Zhang, Wenwei and Chen, Qi Alfred and Liu, Ziwei and Pan, Liang},
  booktitle={Proceedings of the IEEE/CVF International Conference on Computer Vision},
  pages={6585--6597},
  year={2025}
}

@article{caesar2021nuplan,
  title={nuplan: A closed-loop ml-based planning benchmark for autonomous vehicles},
  author={Caesar, Holger and Kabzan, Juraj and Tan, Kok Seang and Fong, Whye Kit and Wolff, Eric and Lang, Alex and Fletcher, Luke and Beijbom, Oscar and Omari, Sammy},
  journal={arXiv preprint arXiv:2106.11810},
  year={2021}
}

@article{tian2024tokenize,
  title={Tokenize the world into object-level knowledge to address long-tail events in autonomous driving},
  author={Tian, Ran and Li, Boyi and Weng, Xinshuo and Chen, Yuxiao and Schmerling, Edward and Wang, Yue and Ivanovic, Boris and Pavone, Marco},
  journal={arXiv preprint arXiv:2407.00959},
  year={2024}
}

@article{swerdlow2024street,
  title={Street-view image generation from a bird's-eye view layout},
  author={Swerdlow, Alexander and Xu, Runsheng and Zhou, Bolei},
  journal={IEEE Robotics and Automation Letters},
  volume={9},
  number={4},
  pages={3578--3585},
  year={2024},
  publisher={IEEE}
}

@article{yang2023bevcontrol,
  title={Bevcontrol: Accurately controlling street-view elements with multi-perspective consistency via bev sketch layout},
  author={Yang, Kairui and Ma, Enhui and Peng, Jibin and Guo, Qing and Lin, Di and Yu, Kaicheng},
  journal={arXiv preprint arXiv:2308.01661},
  year={2023}
}

@inproceedings{li2024drivingdiffusion,
  title={DrivingDiffusion: Layout-guided multi-view driving scenarios video generation with latent diffusion model},
  author={Li, Xiaofan and Zhang, Yifu and Ye, Xiaoqing},
  booktitle={European Conference on Computer Vision},
  pages={469--485},
  year={2024},
  organization={Springer}
}

@inproceedings{wen2024panacea,
  title={Panacea: Panoramic and controllable video generation for autonomous driving},
  author={Wen, Yuqing and Zhao, Yucheng and Liu, Yingfei and Jia, Fan and Wang, Yanhui and Luo, Chong and Zhang, Chi and Wang, Tiancai and Sun, Xiaoyan and Zhang, Xiangyu},
  booktitle={Proceedings of the IEEE/CVF Conference on Computer Vision and Pattern Recognition},
  pages={6902--6912},
  year={2024}
}

@inproceedings{lu2024wovogen,
  title={Wovogen: World volume-aware diffusion for controllable multi-camera driving scene generation},
  author={Lu, Jiachen and Huang, Ze and Yang, Zeyu and Zhang, Jiahui and Zhang, Li},
  booktitle={European conference on computer vision},
  pages={329--345},
  year={2024},
  organization={Springer}
}

@article{gao2024vista,
  title={Vista: A generalizable driving world model with high fidelity and versatile controllability},
  author={Gao, Shenyuan and Yang, Jiazhi and Chen, Li and Chitta, Kashyap and Qiu, Yihang and Geiger, Andreas and Zhang, Jun and Li, Hongyang},
  journal={Advances in Neural Information Processing Systems},
  volume={37},
  pages={91560--91596},
  year={2024}
}

@article{wang2025longdwm,
  title={LongDWM: Cross-granularity distillation for building a long-term driving world model},
  author={Wang, Xiaodong and Wu, Zhirong and Peng, Peixi},
  journal={arXiv preprint arXiv:2506.01546},
  year={2025}
}

@inproceedings{gao2025magicdrive,
  title={MagicDrive-V2: High-resolution long video generation for autonomous driving with adaptive control},
  author={Gao, Ruiyuan and Chen, Kai and Xiao, Bo and Hong, Lanqing and Li, Zhenguo and Xu, Qiang},
  booktitle={Proceedings of the IEEE/CVF International Conference on Computer Vision},
  pages={28135--28144},
  year={2025}
}

@article{hu2024drivingworld,
  title={DrivingWorld: Constructing world model for autonomous driving via video GPT},
  author={Hu, Xiaotao and Yin, Wei and Jia, Mingkai and Deng, Junyuan and Guo, Xiaoyang and Zhang, Qian and Long, Xiaoxiao and Tan, Ping},
  journal={arXiv preprint arXiv:2412.19505},
  year={2024}
}

@inproceedings{fu2024drivegenvlm,
  title={Drivegenvlm: Real-world video generation for vision language model based autonomous driving},
  author={Fu, Yongjie and Jain, Anmol and Chen, Xu and Mo, Zhaobin and Di, Xuan},
  booktitle={2024 IEEE International automated vehicle validation conference (IAVVC)},
  pages={1--6},
  year={2024},
  organization={IEEE}
}

@article{yang2025geniedrive,
  title={GenieDrive: Towards Physics-Aware Driving World Model with 4D Occupancy Guided Video Generation},
  author={Yang, Zhenya and Liu, Zhe and Lu, Yuxiang and Hou, Liping and Miao, Chenxuan and Peng, Siyi and Feng, Bailan and Bai, Xiang and Zhao, Hengshuang},
  journal={arXiv preprint arXiv:2512.12751},
  year={2025}
}

@inproceedings{lu2025infinicube,
  title={Infinicube: Unbounded and controllable dynamic 3d driving scene generation with world-guided video models},
  author={Lu, Yifan and Ren, Xuanchi and Yang, Jiawei and Shen, Tianchang and Wu, Zhangjie and Gao, Jun and Wang, Yue and Chen, Siheng and Chen, Mike and Fidler, Sanja and others},
  booktitle={Proceedings of the IEEE/CVF International Conference on Computer Vision},
  pages={27272--27283},
  year={2025}
}

@article{chen2024unimlvg,
  title={Unimlvg: Unified framework for multi-view long video generation with comprehensive control capabilities for autonomous driving},
  author={Chen, Rui and Wu, Zehuan and Liu, Yichen and Guo, Yuxin and Ni, Jingcheng and Xia, Haifeng and Xia, Siyu},
  journal={arXiv preprint arXiv:2412.04842},
  year={2024}
}

@article{lu2024seeing,
  title={Seeing beyond views: Multi-view driving scene video generation with holistic attention},
  author={Lu, Hannan and Wu, Xiaohe and Wang, Shudong and Qin, Xiameng and Zhang, Xinyu and Han, Junyu and Zuo, Wangmeng and Tao, Ji},
  journal={arXiv preprint arXiv:2412.03520},
  year={2024}
}

@article{yao2024mygo,
  title={Mygo: Consistent and controllable multi-view driving video generation with camera control},
  author={Yao, Yining and Guo, Xi and Ding, Chenjing and Wu, Wei},
  journal={arXiv preprint arXiv:2409.06189},
  year={2024}
}

@inproceedings{guo2025dist,
  title={Dist-4d: Disentangled spatiotemporal diffusion with metric depth for 4d driving scene generation},
  author={Guo, Jiazhe and Ding, Yikang and Chen, Xiwu and Chen, Shuo and Li, Bohan and Zou, Yingshuang and Lyu, Xiaoyang and Tan, Feiyang and Qi, Xiaojuan and Li, Zhiheng and others},
  booktitle={Proceedings of the IEEE/CVF International Conference on Computer Vision},
  pages={27231--27241},
  year={2025}
}

@article{guo2024infinitydrive,
  title={Infinitydrive: Breaking time limits in driving world models},
  author={Guo, Xi and Ding, Chenjing and Dou, Haoxuan and Zhang, Xin and Tang, Weixuan and Wu, Wei},
  journal={arXiv preprint arXiv:2412.01522},
  year={2024}
}

@article{sun2021loftr,
  title={{LoFTR}: Detector-Free Local Feature Matching with Transformers},
  author={Sun, Jiaming and Shen, Zehong and Wang, Yuang and Bao, Hujun and Zhou, Xiaowei},
  journal={CVPR},
  year={2021}
}

@article{hu2022lora,
  title={Lora: Low-rank adaptation of large language models.},
  author={Hu, Edward J and Shen, Yelong and Wallis, Phillip and Allen-Zhu, Zeyuan and Li, Yuanzhi and Wang, Shean and Wang, Liang and Chen, Weizhu and others},
  journal={Iclr},
  volume={1},
  number={2},
  pages={3},
  year={2022}
}

@article{li2026far,
  title={FAR-Drive: Frame-AutoRegressive Video Generation in Closed-Loop Autonomous Driving},
  author={Li, Yaoru and Landi, Federico and Godi, Marco and Jin, Xin and Fu, Ruiju and Ma, Yufei and Sun, Muyang and Si, Heyu and Guo, Qi},
  journal={arXiv preprint arXiv:2603.14938},
  year={2026}
}

@inproceedings{jiang2025vace,
  title={Vace: All-in-one video creation and editing},
  author={Jiang, Zeyinzi and Han, Zhen and Mao, Chaojie and Zhang, Jingfeng and Pan, Yulin and Liu, Yu},
  booktitle={Proceedings of the IEEE/CVF International Conference on Computer Vision},
  pages={17191--17202},
  year={2025}
}

@article{kerbl20233d,
  title={3d gaussian splatting for real-time radiance field rendering.},
  author={Kerbl, Bernhard and Kopanas, Georgios and Leimk{\"u}hler, Thomas and Drettakis, George and others},
  journal={ACM Trans. Graph.},
  volume={42},
  number={4},
  pages={139--1},
  year={2023}
}
